\documentclass[journal]{IEEEtran}
\usepackage{amsmath}
\usepackage{booktabs} 
\usepackage{cases}
\usepackage{subfigure}
\usepackage{graphicx}
\usepackage{CJK}
\usepackage{color}
\usepackage{multirow}
\usepackage[T1]{fontenc}
\newcommand{\eat}[1]{}
\begin{document}
%
\title{Self-Supervised Video Hashing with Hierarchical Binary Auto-encoder}
%
%
%

\author{Jingkuan Song,
        Hanwang Zhang, Xiangpeng Li, Lianli Gao, Meng Wang and Richang Hong
\thanks{Jingkuan Song, Xiangpeng Li and Lianli Gao are with the Center of Future Media, School of Computer Science and Engineering, University of Electronic Science and Technology of China, 611731. Hanwang Zhang is from Nanyang University of Technology, Singapore. Meng Wang and Richang Hong are with Hefei University of Technology, Hefei, China.
}
\thanks{Manuscript received August 19, 2017.}}

%
%

\markboth{Transactions on Image Processing}%
{Shell \MakeLowercase{\textit{et al.}}: Bare Demo of IEEEtran.cls for Journals}
%



\maketitle

\begin{abstract}

    Existing video hash functions are built on three isolated stages: frame pooling, relaxed learning, and binarization, which have not adequately explored the temporal order of video frames in a joint binary optimization model, resulting in severe information loss. 
    In this paper, we propose a novel unsupervised video hashing framework dubbed Self-Supervised Video Hashing (SSVH), that is able to capture the temporal nature of videos in an end-to-end learning-to-hash fashion. We specifically address two central problems: 1) how to design an encoder-decoder architecture to generate binary codes for videos; and 2) how to equip the binary codes with the ability of accurate video retrieval. We design a hierarchical binary autoencoder to model the temporal dependencies in videos with multiple granularities, and embed the videos into binary codes with less computations than the stacked architecture. Then, we encourage the binary codes to simultaneously reconstruct the visual content and neighborhood structure of the videos. Experiments on two real-world datasets (FCVID and YFCC) show that our SSVH method can significantly outperform the state-of-the-art methods and achieve the currently best performance on the task of unsupervised video retrieval.

   \eat{  SSTH novelly proposed an encoder-decoder frame using Binary LSTM (BLSTM) to encode videos into binary codes which solve these problems to some extent. But the learned hash codes still cann't represent long dependencies of video well. So we further extend our model to address these issues and propose a Self-Supervised Video Hashing (SSVH). SSVH follow BLSTM and encoder-decoder framework of SSTH. Besides, we first design a hierarchical binary auto-encoder to model the temporal dependencies in videos with multiple granularities, and embed the videos into binary codes with less computations than the stacked architecture. Then, we encourage the binary codes to simultaneously reconstruct the visual content and neighborhood structures of the videos. Experiments on two real-world datasets (FCVID and YFCC) show that our SSVH method can significantly outperform the state-of-the-art methods and achieve the currently best performance on the task of unsupervised video retrieval.}

\end{abstract}


%
\IEEEpeerreviewmaketitle

\section{Introduction}

\label{sec.intro}
Nowadays, due to the advances in transmission technologies, capture devices and display techniques, we are witnessing the rapid growth of videos and video retrieval related services. Take YouTube and Facebook as examples. According to Youtube Statistics 2017, 300 hours of videos are uploaded to Youtube every minute, and it attracts over 30 million visitors per day. For Facebook, it also has 8 billion average daily video views from 500 million users. Therefore, the explosive growth of online videos makes large-scale content-based video retrieval (CBVR) \cite{zhang2016play,ye2013large,song2011multiple} an urgent need. However, unlike Content-based Image Retrieval (CBIR) that has been extensively studied in the past decades \cite{song2017deep,DBLP:journals/pami/GongLGP13,LiuWJJC12} and considerable progress has been achieved, CBVR  has not received sufficient attention in multimedia community~\cite{smeulders2000content,datta2008image,wang2012semi}. 

\eat{However, video hashing has not yet received sufficient attention. On the other hand, the multimedia community has been working on content-based image retrieval (CBIR) \cite{song2017deep,DBLP:journals/pami/GongLGP13,LiuWJJC12} for decades and considerable progress has been achieved.  }

On the other hand, hashing methods have been widely acknowledged as a good solution for approximate nearest neighbor search by transforming high-dimensional features as short binary codes to support efficient large-scale retrieval and data storage.
Therefore, content-based video hashing is a promising direction for CBVR, but video is beyond a set of frames and video retrieval is more challenging than image hashing. Essentially, the rich temporal information in videos is a barrier for directly utilizing image-based hashing methods.  However, most current works on video analytics generally resort to pooling frame-level features into a single video-level feature by discarding the temporal order of the frame sequence \cite{ye2013large,song2011multiple,cao2012submodular}. Such bag-of-frames degeneration works well when high-dimensional frame-level features such as CNN responses \cite{wu2016harnessing} and motion trajectories \cite{wang2013action} are used, as certain temporal information encoded in a high dimension can be preserved after pooling. However, for large-scale CBVR, where hashing (or indexing) of these high-dimensional features as short binary codes is necessary, the temporal information loss caused by frame pooling will inevitably result in suboptimal binary codes of videos. The loss usually takes place in the process of hash function learning \cite{wang2016learning}, which is a post-step after pooling; Compared to dominant video appearances (e.g., objects, scenes and short-term motions), nuanced video dynamics (e.g., long-term event evolution) are more likely to be discarded as noise in the drastic feature dimensionality reduction during hashing \cite{donahue2015long}.

\eat{
A few attempts take a video as a set of images and ignore the temporal information \cite{ye2013large,song2011multiple,cao2012submodular}. Although the performance is promising, these methods learn hash codes for each frame where the structure of a video is not introduced.}

\eat{Although pooling provides a solution for video hashing generation, it inevitably results in suboptimal binary codes, since video temporal information is significantly ignored. In general, they firstly conduct pooling on frame-level deep features to generate video-level features, and then project this high-dimensional features into low-dimensional hash codes \cite{liong2016deep}, thus video temporal information is significantly ignored.}

Recently, deep learning has dramatically improved the state-of-the-art in speech recognition, visual object detection and image classification \cite{krizhevsky2012imagenet,girshick2015fast,you2016image}. Inspired by this, researchers \cite{liong2016deep} started to extract from various deep Convolutional Neural Networks (Deep ConvNets) (e.g., VGG \cite{simonyan2014very} and RestNet \cite{he2016deep}) to support video hashing, but video temporal information is significantly ignored. To capture the temporal information, the Recurrent Neural Network (RNN) is used to achieve the state-of-the-art performance in sequential data streams {\cite{venugopalan2015sequence,pan2016hierarchical}}. RNN-based hashing methods \cite{gu2016supervised} utilized RNN and video labels to learn discriminative hash codes for videos. However, human labels are time- and labor- consuming, especially for large-scale datasets. Therefore, we argue that the key reason to the above defect is that both the temporal pooling and the hash code learning steps have not adequately addressed the temporal nature of videos. Also,  we argue that the hash codes which can simply reconstruct the video content are unable to achieve high accuracy for the task of video retrieval.

Self-Supervised Temporal Hashing (SSTH) \cite{zhang2016play} aims to improve video hashing by taking temporal information into consideration. A stacking strategy is proposed to integrate Long Short-Term Memory Networks (LSTMs) \cite{hochreiter1997long} with hashing to simultaneously preserve the temporal and spatial information of a video using hash codes. Despite the improved performance, a major disadvantage of stacked LSTM is that it introduces a long path from the input to the output video vector representation, therefore may result in heavier computational cost.

Therefore, to address the issues of leveraging temporal information and reducing computational as costs as mentioned above, we improve the version of SSTH from \cite{zhang2016play} by  proposing  Self-Supervised Video Hashing (SSVH), which encodes video temporal and visual information simultaneously using an end-to-end hierarchical binary auto-encoder and a neighborhood structure. It is worth to highlight the following contributions.

\eat{More recently, Zhang \textit{et al.} proposed a Self-Supervised Temporal Hashing (SSTH) \cite{zhang2016play}, which aims to improve video hashing by taking temporal information into consideration. They proposed a stacking strategy to integrate Long Short-Term Memory Networks (LSTMs) \cite{hochreiter1997long} with hashing to simultaneously preserving a video's original temporal and spatial information using a short hash code. Despite the improved performance, a major disadvantage of stacked LSTM is that it introduces a long path from the input to the output video vector representation, therefore may result in heavier computational cost. Also, we argue that the hash codes which can simply reconstruct the video content are unable to achieve high accuracy for the task of video retrieval.
}

\begin{itemize}
	\eat{\item We propose an end-to-end deep learning framework named Self-Supervised Video Hashing (SSVH) to generate compact binary codes from videos. It is a generic framework, which can be easily extend to other types of sequence data streams such as sentences, speeches, etc.}
	\item { We develop a novel LSTM variant dubbed Binary LSTM (BLSTM), which serves as the building block of the temporal-aware hash function. We also develop an efficient backpropagation rule that directly tackles the challenging problem of binary optimization for BLSTM without any relaxation.}
	\item We design a hierarchical binary auto-encoder to model the temporal dependencies in videos with multiple granularities, and embed the videos into binary codes with less computations. To achieve accurate video retrieval, we impose the binary codes to simultaneously reconstruct the visual content and neighborhood structures of the videos.	 	
	\item We successfully show that our SSVH model achieves promising results on two large-scale video dataset FCVID ($91,223$ videos) and YFCC ($700,882$ videos). Experimental results on two datasets show superior performance compared to other unsupervised video hashing methods.
\end{itemize}

{Compared with SSTH, we use hierarchical structure to replace stacked LSTMs which can efficiently learn long-range dependency and reduce computational cost. Besides, we also use a neighborhood structure to enhance the representation ability of binary codes. Experimental results verify the effectiveness of our proposed method.}

\begin{figure*}[t]
	\centering
	\includegraphics[width=1.0\textwidth]{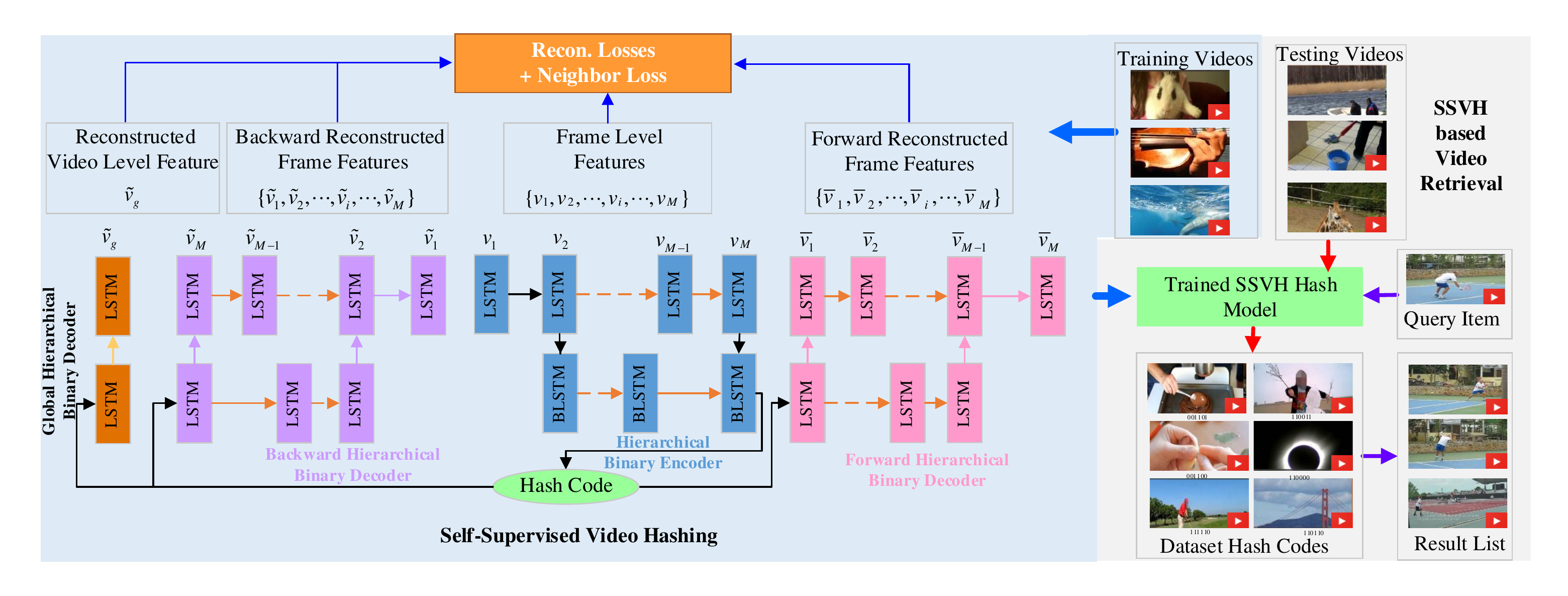}
	\centering
	\caption{The overview of the proposed Self-Supervised Video Hashing for content based video retrieval. The left part with light blue background denotes the off-line training with reconstruction losses and a neighborhood structure loss. The right part with the gray background denotes the content-based video retrieval process. }
	\label{fig.framework}
\end{figure*}

\section{Related work}
\label{sec.relatedwork}
In this section, we describe related work. Our SSVH is closely related to hashing in terms of the algorithm, and video content analysis in terms of the application.
 
\subsection{Hashing}

The rapid growth of massive databases in various applications has promoted the research and study of hash-based indexing algorithm. Learning to hash \cite{wang2017survey} has been widely applied to approximate nearest neighbor search for large-scale multimedia data, due to its computation efficiency and retrieval quality. 
Learning-based hashing learns a hash function, $\mathbf{y} = \mathbf{h}(\mathbf{x}) \in \{0,1\}^L$, mapping an input item $\mathbf{x}$ to a compact code $\mathbf{y}$.
By mapping data into binary codes, efficient storage and search can be achieved due to the fast bit XOR operations in Hamming space. Current hashing methods can be generally categorized into supervised and unsupervised ones. 

Supervised hashing methods \cite{gu2016supervised,ye2013large,liong2016deep,li2015feature,LiuWJJC12} are proposed to utilize available supervision information like class labels or pairwise labels of the training data for improving the performance of hash codes. 
Ye \textit{et. al.} \cite{ye2013large} proposed a supervised framework Video Hashing with both Discriminative commonality and Temporal consistency (VHDT) with structure learning to design efficient linear hash functions for video indexing and formulate the objective function as a minimization problem of a structure-regularized empirical loss function. But it only generated frame-level codes.
Cao \textit{et. al.}~\cite{cao2017hashnet} proposed a novel deep learning architecture named HashNet to learn hash codes by a continuation method, which learned exactly binary codes from imbalanced similarity labels but ignored the temporal information of video frames.
Deep Pairwise-Supervised Hashing (DPSH) \cite{li2015feature} took into account the pairwise relationship and proposed a novel deep hashing method to perform simultaneous features extraction and hash codes learning. Supervised Recurrent Hashing (SRH) \cite{gu2016supervised} was proposed to deploy the Long Short-Term Memory (LSTM) to model the structure of video samples and introduce a max-pooling mechanism to embedding the frames into fixed-length representations that are fed into supervised hashing loss. In addition, Liong \textit{et.al}~\cite{liong2016deep} proposed a method named Deep Video Hashing (DVH), which utilized spatial-temporal information after the stacked convolutional pooling layers to extract representative video features, and then obtained compact binary codes.

Unsupervised hashing methods \cite{DBLP:journals/pami/GongLGP13,song2017deep,erin2015deep,zhang2016play,song2011multiple,song2018quantization} often utilize the data properties such as distribution or manifold structure to design effective indexing schemes. 
ITQ \cite{DBLP:journals/pami/GongLGP13} rotated data points iteratively to minimize the quantization loss. Liong \textit{et.al} \cite{erin2015deep} proposed to use deep neural network to learn hash codes by three objective: (1) the loss between the real-valued feature descriptor and the learned binary codes, (2) binary codes distribute evenly on each bit and (3) different bits are as independent as possible. Most of the proposed methods are devoted to image retrieval, which cannot be directly applied to video hashing due to its inherent temporal information. A more completed survey of hashing methods can be found in~\cite{wang2016survey}.

{There are also some research focusing on video hashing. For example, the hash functions proposed by Song et al. \cite{song2011multiple} and Cao et al.~\cite{cao2017hashnet} are unaware of the temporal order of video frames. Ye et al. \cite{ye2013large} exploited the pairwise frame order but their method requires video labels and only generated frame-level codes. Although Revaud et al. \cite{revaud2013event} exploited the short-term temporal order, their video quantization codes were not binary. In contrast to the above methods, our SSVH is an unsupervised binary code learning framework that explicitly exploits the long-term video temporal information.}

\eat{
Song \textit{et. al.} \cite{song2011multiple} is limited with the temporal nature of videos and is unaware of the temporal order of video frames.
Zhang \textit{et. al.} \cite{zhang2016play} proposed a video hashing method named SSTH to encode the video features to binary code and then use a decoder to reconstruct the original video frames from the binary codes. SSTH reconstructs the video features in forward and reverse order, and the temporal nature of videos is captured in an end-to-end learning-to-hash fashion.	}

\subsection{Video Content Analysis based on LSTM}

A video is a sequence of frames. But it is beyond a set of frames and the temporal information in video is important for video content analysis. 
To capture the temporal order, Ng \textit{et al.} \cite{yue2015beyond} introduced Long Short-Term Memory (LSTM) inspired by the general sequence to sequence learning neural model proposed by Sutskever \textit{et al.} \cite{sutskever2014sequence}. 
Since then, LSTM has been widely used in different research fields such as image and video captioning \cite{you2016image,venugopalan2015sequence,song2017hierarchical}, nature language processing \cite{li2017adversarial,britz2017massive} and visual question-answering \cite{kafle2016answer,yang2016stacked}. 
The basic video content analysis model is an encoder-decoder framework composed of LSTM. They usually use deep neural network as an encoder to extract frame-level features.
Vinyals \textit{et al.}~\cite{VinyalsTBE15} proposed to use LSTM to decode the latent semantic information of images and generate words in order. 
Sequence to sequence model \cite{venugopalan2015sequence} used a two-layer stacked LSTM to encode and decode the static and motion information to generate the video caption. 
Hierarchical Recurrent Neural Encoder \cite{pan2016hierarchical} was able to exploit video temporal structure in a longer range by reducing the length of input information flow. Hierarchical LSTM layer was able to uncover the temporal transitions between frame chucks with different granularities and can model the temporal transpositions between frames as well as the transitions between segments. 
Different from the stacked LSTM which simply aims to introduce more non-linearity into the neural model, the Hierarchical LSTM aimed to abstract the visual information at different time scales, and learn the visual features with multiple granularities.
Song \textit{et al.} \cite{song2017hierarchical} proposed a hierarchical LSTM with adaptive attention to generate caption. This model used LSTM to encode video content and semantic information, and decode video information by adaptive attention to generate video caption.
SAN (Stacked Attention Network) were proposed with CNN and RNN as an encoder to explore visual structure. 
Encoder-decoder framework composed of LSTM is popular in sequential data understanding and makes a great success. We also choose LSTM as a basic unit in our encoder-decoder model.

\section{Proposed Method}
\label{sec.main1}
In this section, we formulate the proposed Self-Supervised Video Hashing framework (Fig.~\ref{fig.framework}). Notations and problem definition that will be used in the rest of the paper will be introduced first. Then we will present two major components of our novel framework, Hierarchical Binary Auto-Encoder and Neighborhood Structure. Finally, optimization method is introduced to train our model.

\subsection{Notations and Problem Definition}
Given a video $\mathbf{V}=[\mathbf{v}_1,...,\mathbf{v}_M] \in {R}^{M \times D}$, where $M$ is the number of frames in each video\footnote{For each video, we extract equal number of frames}, and $D$ is the feature dimensionality of each frame. $N$ denotes the number of videos in  dataset. The features for each frame in a video are extracted as a pre-processing step, and $\mathbf{v_i}$ indicates the $i$-th frame features of a video. The goal of a video hashing method is to learn a binary code $\mathbf{b} \in \{-1,1\} ^L$ for each video where $L$ is the code length. $\mathbf{h}$ is the hidden state of BLSTM unit before $sgn$ function, and $\mathbf{b} = sgn(\mathbf{h})$.

	\subsection{Hierarchical Binary Auto-Encoder}
	\label{subsec.hbae}
	
	Long Short-Term Memory (LSTM) is popular in video analysis in recent years because of the effectiveness of processing sequential information. However, the original LSTM can only generate continuous values instead of binary codes. A straightforward way is to add a hashing layer which consists of a full connected layer to obtain the hidden variable $\textbf{h}$, and a $sgn$ activation layer to binarize the $\textbf{h}$ to binary codes $\textbf{b}$. However, as pointed out in \cite{zhang2016play}, this strategy is essentially based on frame pooling, where the pooling function is an RNN. Even though the pooling is temporal-aware, the hash codes do not directly capture the temporal nature of videos.

	In order to design an architecture that can simultaneously capture the temporal information of videos and generate binary codes, we propose a novel hierarchical binary auto-encoder based on hierarchical LSTM \cite{pan2016hierarchical}. Specifically, our architecture consists of {an encoder and a decoder}, while the decoder is divided into forward hierarchical binary decoder, backward hierarchical binary decoder and global hierarchical binary decoder.
	
	\begin{figure}[t]
		\centering
		\includegraphics[width=0.9\linewidth]{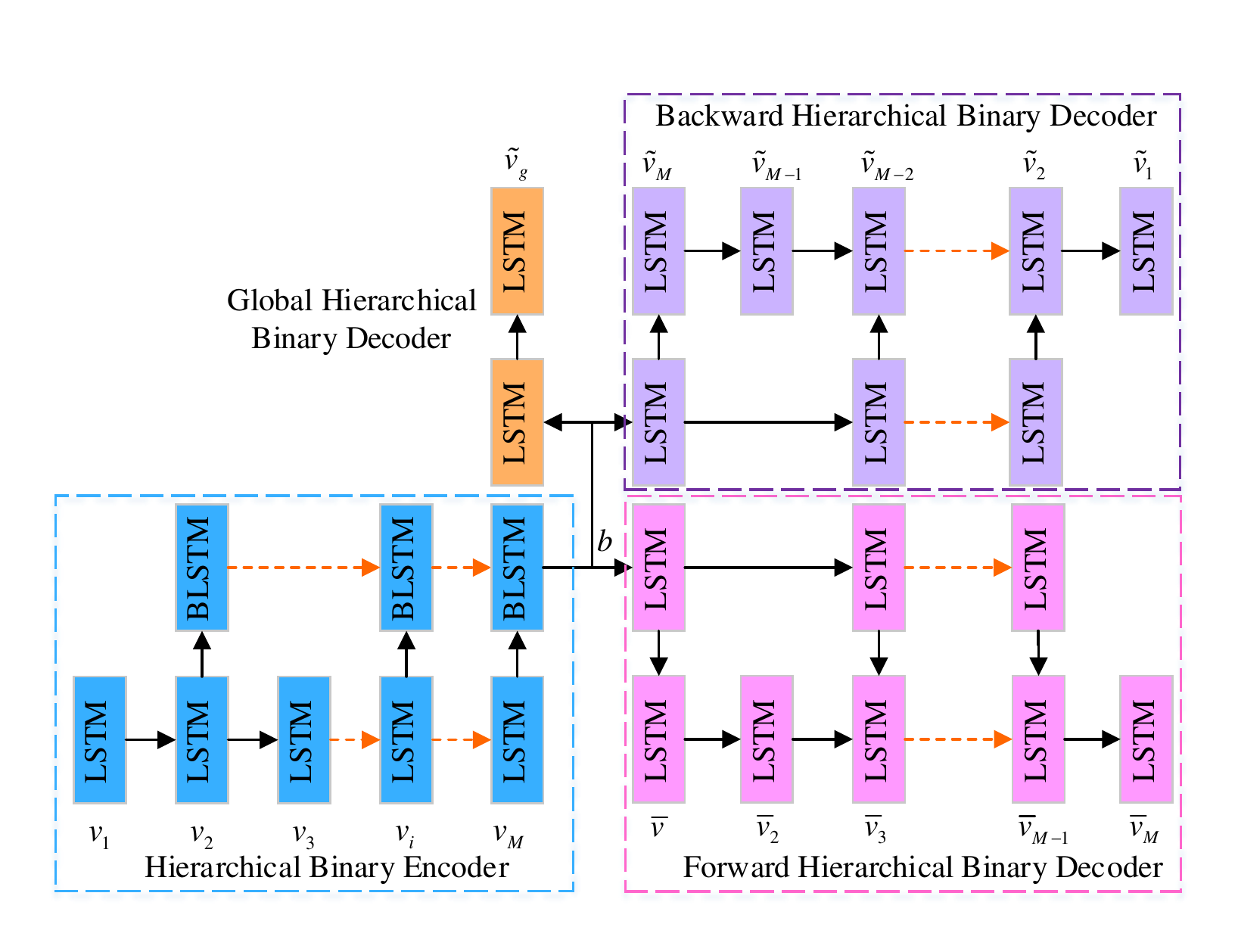}
		\centering
		\caption{The overview of encoder-decoder. The input video $V=[v_1,...,v_M]$ is firstly mapped to binary codes by the Hierarchical Binary Encoder. Then, we reconstruct the video using Forward Hierarchical Binary Decoder, Backward Hierarchical Binary Decoder and Global Hierarchical Binary Decoder.}
		\label{fig.encoder}
	\end{figure}

	\begin{figure}[t]
		\centering
		\includegraphics[width=0.7\linewidth]{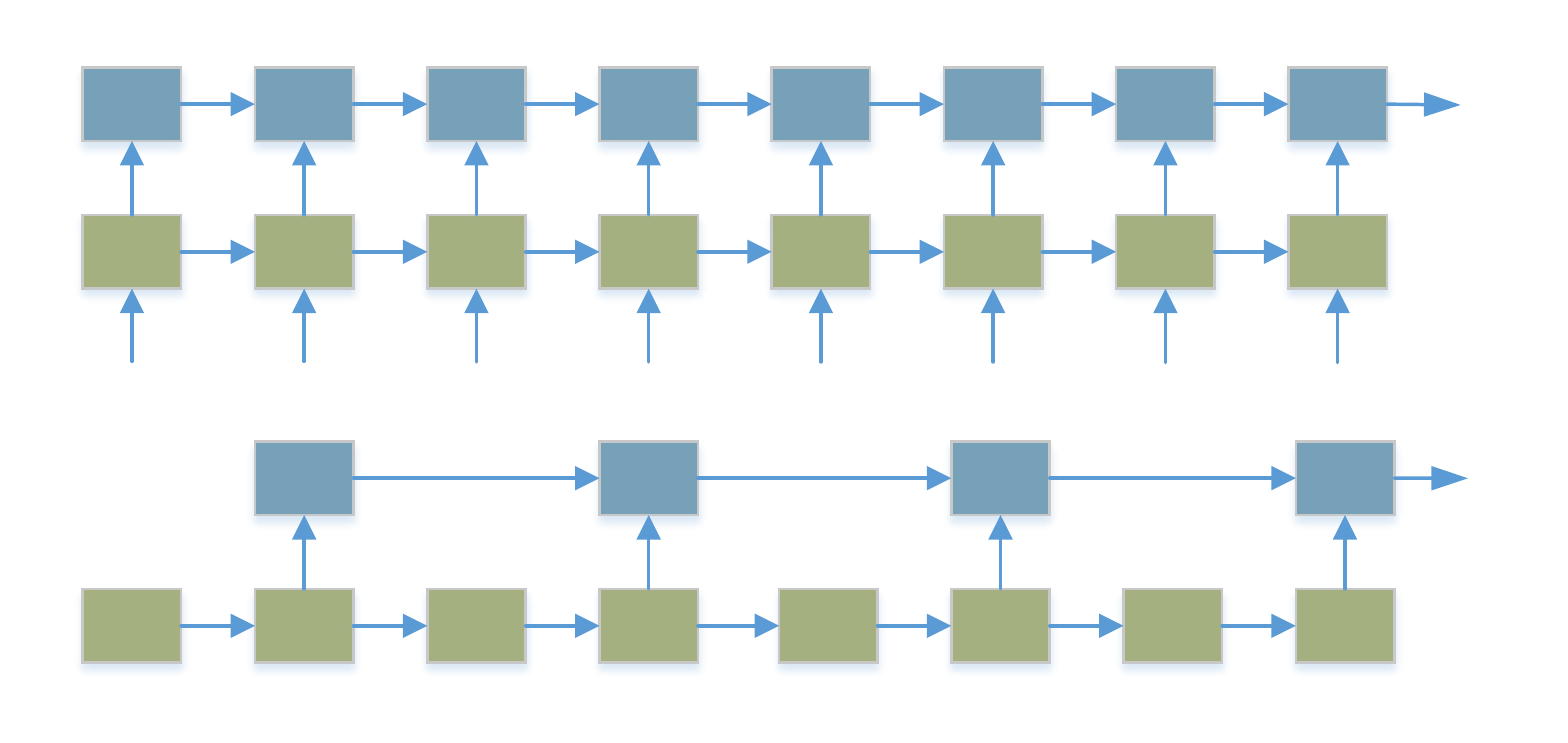}
		\centering
		\caption{{The difference between a stacked LSTM and a hierarchical LSTM. The above framework means a stacked architecture and the below framework is a hierarchical structure (with stride=2). The basic unit usually denotes a RNN unit (LSTM or GRU) in the encoder-decoder framework.}}
		\label{fig.hier_stack}
	\end{figure}
	
	\subsubsection{Hierarchical Binary Encoder}
	
	As can be seen in Fig.~\ref{fig.encoder}, the binary encoder is a two-layered hierarchical RNN structure, and it consists of vanilla LSTM and binary-LSTM. 
	The first layer is an original LSTM which can be considered as a higher-level feature extractor for the frames. The hidden state at time step $i$ is $\mathbf{z_i}$:
	\begin{align}
	\mathbf{z}_i = f(\mathbf{v}_i, \mathbf{z}_{i-1})
	\end{align}
	where $\mathbf{v}_i$ indicates the features of the $i$-th frame, $\mathbf{z}_{i-1}$ is the hidden state at time step $i-1$ ($\mathbf{z}_0=0$) and f($\cdot$) is the function of this LSTM layer. The hidden state is used as input to the second layer (BLSTM).
	
	The second layer is a binary LSTM layer (Fig.~\ref{fig.encoder}), which embeds the higher-level real-valued features of a video to a binary code. To achieve this, a straightforward way is stacking another layer, as introduced in \cite{zhang2016play}. However, it will increase computation operations. Inspired by \cite{pan2016hierarchical}, we proposed a hierarchical binary LSTM.
	Different from stacked LSTM \cite{pan2016hierarchical}, in the hierarchical binary LSTM, not all the output of the first-layer LSTM is connected to the second-layer BLSTM. 
	The motivation is that this hierarchical structure can efficiently exploit video temporal structure in a longer range by reducing the length of input information flow, and compositing multiple consecutive frames at a higher level. Also, computation operations are significantly reduced. The difference between a stacked LSTM and a hierarchical LSTM is illustrated in Fig.\ref{fig.hier_stack}. 
	
	Suppose the stride of the hierarchical BLSTM is $l$. Then, only the output $\mathbf{z}_{l\times i}$ of the $(l\times i)$-th time step in the first layer LSTM will be used as input to $i$-th time step in the second BLSTM layer.
	Our BLSTM follows a similar data flow as LSTM, and the detailed implementation of BLSTM is given as follows:
	\begin{eqnarray}
	\mathbf{f}_t = \sigma (\mathbf{W}_{zf}\mathbf{z}_t + \mathbf{U}_{bf}\mathbf{b}_{t - 1} + \mathbf{M}_{cf} \circ \mathbf{c}_{t - 1} + \mathbf{b}_f)\\
	\mathbf{i}_t = \sigma (\mathbf{W}_{zi}\mathbf{z}_t + \mathbf{U}_{bi}\mathbf{b}_{t - 1} + \mathbf{M}_{ci} \circ \mathbf{c}_{t - 1} + \mathbf{b}_i)\\
	\mathbf{o}_t = \sigma (\mathbf{W}_{zo}\mathbf{z}_t + \mathbf{U}_{bo}\mathbf{b}_{t - 1} + \mathbf{M}_{co} \circ \mathbf{c}_{t - 1} + \mathbf{b}_o)\\
	\mathbf{m}_t = \phi (\mathbf{W}_{zm}\mathbf{z}_t + \mathbf{U}_{bm}mathbf{b}_{t - 1} + \mathbf{b}_m)\\
	\mathbf{c}_t = batch\_norm(\mathbf{f}_t \circ \mathbf{c}_{t - 1} +\mathbf{i}_t \circ \mathbf{m}_t)\\
	\mathbf{h}_t = \mathbf{o}_t \circ \mathbf{c}_t\\
	\mathbf{b}_t = {\mathop{\rm sgn}} (\mathbf{h}_t)
	\label{equ.sgn}
	\end{eqnarray}
	where $ \circ $ denotes the element-wise multiplication and $batch\_norm$ means batch normalization. Therefore, the output of our encoder will be a binary code $\mathbf{b} \in \{-1,1\} ^L$. The behaviors of ``forget'', ``input'' and ``output'' in BSLTM unit are respectively controlled by three gate variables: forget gate $\mathbf{f}_t$, input gate $\mathbf{i}_t$, and output gate $\mathbf{o}_t$. $\mathbf{W}$, $\mathbf{U}$ and $\mathbf{M}$ denote the shared weight matrices of BLSTM to be learned and $\mathbf{b}$ means bias term. $\mathbf{m}_t$ is the input to the memory cell $\mathbf{c}_t$, which is gated by the input gate $\mathbf{i}_t$. $\sigma$ denotes the element-wise logistic sigmoid function and $\phi$ denotes hyperbolic tangent function tanh.
	
	\subsubsection{Forward Hierarchical Binary Decoder}
	The forward hierarchical binary decoder reconstructs the input frame features in a forward order $\bar{\mathbf{v}}_1,\bar{\mathbf{v}}_2,...,\bar{\mathbf{v}}_M$ using the binary codes $\mathbf{b}$. 
	The decoder also has a hierarchical structure which consists of two layers of LSTM. Specifically, the hidden state at time step $i$ of the first layer LSTM is $\mathbf{\bar{z}}_i$:
	\begin{align}
	\mathbf{\bar{z}}_i = \bar{f}(\mathbf{\bar{z}}_{i-1}, 0)
	\end{align}
	where $\bar{f}(.)$ is the function for the forward LSTM, $\mathbf{\bar{z}}_{i-1}$ is the hidden state at time step $i-1$ and $\mathbf{\bar{z}_{0}} = b$.
	
	Similarly, the output $\mathbf{\bar{z}}_{i}$ in the first layer LSTM will not be connected to all the units in the second layer LSTM. Suppose we have the same stride of $l$. Then, the output $\mathbf{\bar{z}}_{i}$ of the $i$-th time step in the first layer LSTM will be used as input to $i\times l$-th time step in the second LSTM layer. The reconstructed $\mathbf{\bar{z'}}_j$ is formulated as:
	\begin{align}
	\mathbf{\bar{z'}}_j = \bar{f'}(\mathbf{\bar{z'}}_{j-1}, \mathbf{\bar{z^ \circ}}_j)
	\end{align}
	where $\bar{f'}(.)$ is the function for the second layer of the forward LSTM, $\mathbf{\bar{z'}}_{j-1}$ is the hidden state at time step $j-1$ of the second layer LSTM ($\mathbf{\bar{z'}}_{0} = 0$), $\mathbf{\bar{z^ \circ}}_j=\mathbf{\bar{z}}_{(j-1)/l+1}$ if $j-1$ is a multiple of $l$ and $\mathbf{\bar{z^ \circ}}_j=0$ otherwise.
	
	Then the reconstructed features will be attained by linear reconstructions for the output of the decoder LSTMs:
	\begin{align}
	\mathbf{\bar{v}}_j = \mathbf{\bar{W}} \times \mathbf{\bar{z'}}_{j} +\mathbf{\bar{r}}
	\end{align}
	where $\mathbf{\bar{W}}$ is weight matrix and $\mathbf{\bar{r}}$ means the bias.
	
	We can define the forward decoder loss of a video as the Euclidean distance of the original features and the reconstructed features as:
	\begin{equation}
	Loss_{f} = \sum\limits_{t = 1}^M {||{\mathbf{v}_{t}}-  \mathbf{\bar{v}}_{t}||^2}
	\label{equ.forward}
	\end{equation}

	\subsubsection{Backward Hierarchical Binary Decoder}
	The backward hierarchical binary decoder is similar to the forward hierarchical decoder. It reconstructs frame-level features in a reverse order, i.e., $\mathbf{\tilde{v}}_{M}, ..., \mathbf{\tilde{v}}_{i},\mathbf{\tilde{v}}_{1}$. 
	Specifically, the hidden state at time step $i$ of the first layer LSTM is $\mathbf{\tilde{z}}_i$:
	\begin{align}
	\mathbf{\tilde{z}}_i = \tilde{f}(\mathbf{\tilde{z}}_{i-1}, 0)
	\end{align}
	where $\tilde{f}(.)$ is the function for the backward LSTM, $\mathbf{\tilde{z}}_{i-1}$ is the hidden state at time step $i-1$ and $\mathbf{\bar{z}}_{0} =\mathbf{b}$.
	
	The reconstructed $\mathbf{\tilde{v}_j}$ is formulated as:
	\begin{align}
	\mathbf{\tilde{z'}_j} = \tilde{f'}(\mathbf{\tilde{z'}}_{j+1}, \mathbf{\tilde{z^\circ}}_j)
	\end{align}
	where $\tilde{f'}(.)$ is the function for the second layer of the backward LSTM, $\tilde{\mathbf{z'}}_{j+1}$ is the hidden state at time step $M-j$ of the second layer LSTM ($\mathbf{\tilde{z'}}_{0} = \textbf{0}$), $\mathbf{\tilde{z^\circ}}_j=\mathbf{\tilde{z}}_{(M-j)/l+1}$ if $(M-j)$ is a multiple of $l$ and $\mathbf{\tilde{z^\circ}}_j=\textbf{0}$ otherwise.
	
	Then the reconstructed features will be attained by linear reconstructions for the output of the decoder LSTMs:
	\begin{align}
	\mathbf{\tilde{v}}_j = \tilde{\mathbf{W}} \times \mathbf{\tilde{z'}}_{j} +\tilde{\mathbf{r}}
	\end{align}
	where $\tilde{\mathbf{W}}$ is weight matrix and $\tilde{\mathbf{r}}$ means the bias.
	
	The backward decoder loss is defined as:
	\begin{equation}	
	Loss_{b} = \sum\limits_{t = M}^1 {||\mathbf{v}_{t}-\mathbf{\tilde{v}}_{t}||^2} 
	\label{equ.backward}
	\end{equation}

	\subsubsection{Global Hierarchical Binary Decoder}
	
	Apart from the forward and backward hierarchical binary decoder, we also use a global hierarchical decoder to reconstruct the video level features. Here, we use mean-pooling of all the frame-level features as the video-level feature. The LSTM used in both layers are two basic LSTMs. The global reconstruction loss of a video is defined as:
	\begin{equation}	
	Loss_{g} = ||{\mathbf{v}_{g}}-\mathbf{\tilde{v}}_{g}||^2
	\label{equ.general}
	\end{equation}
	${\mathbf{v}_g} = \frac{1}{{\rm{M}}}\sum\nolimits_{j = 1}^M {{\mathbf{v}_i}}$ indicates the mean of all frame-level features of a video.

	\subsubsection{Hierarchical Binary Auto-Encoder Loss Function} The hierarchical binary auto-encoder consists of three components, and the reconstruction loss is also composed of forward loss, backward loss and global loss, which is defined as:
	\begin{equation}	
	ReconLoss = Loss_f + Loss_b + Loss_g
	\label{equ.recon}
	\end{equation}
	
	\subsection{Neighborhood Structure}
	\label{sub.sec.pairwiseModel}
	
	We argue that achieving a good quality of video content reconstruction is not enough to equip the binary codes with the ability of accurate video retrieval. 
	{ Using basic reconstruction loss, we can learn a binary code of a video, which can only reconstruct the video.}
	Lots of previous studies shows that it is beneficial to exploit the data structure for learning a low-dimensional embedding for the retrieval task.
	{Neighborhood structure enforces similar videos to have close binary codes and dissimilar videos to have different binary code.}
	Inspired by this, we propose a novel method to exploit the neighborhood structure of videos. Then we can train our model to encourage the binary codes to preserve this neighborhood structure.
	
	\subsubsection{Neighborhood Structure Construction}
	
	\begin{figure}[t]
		\centering
		\includegraphics[width=0.9\linewidth,height=5cm]{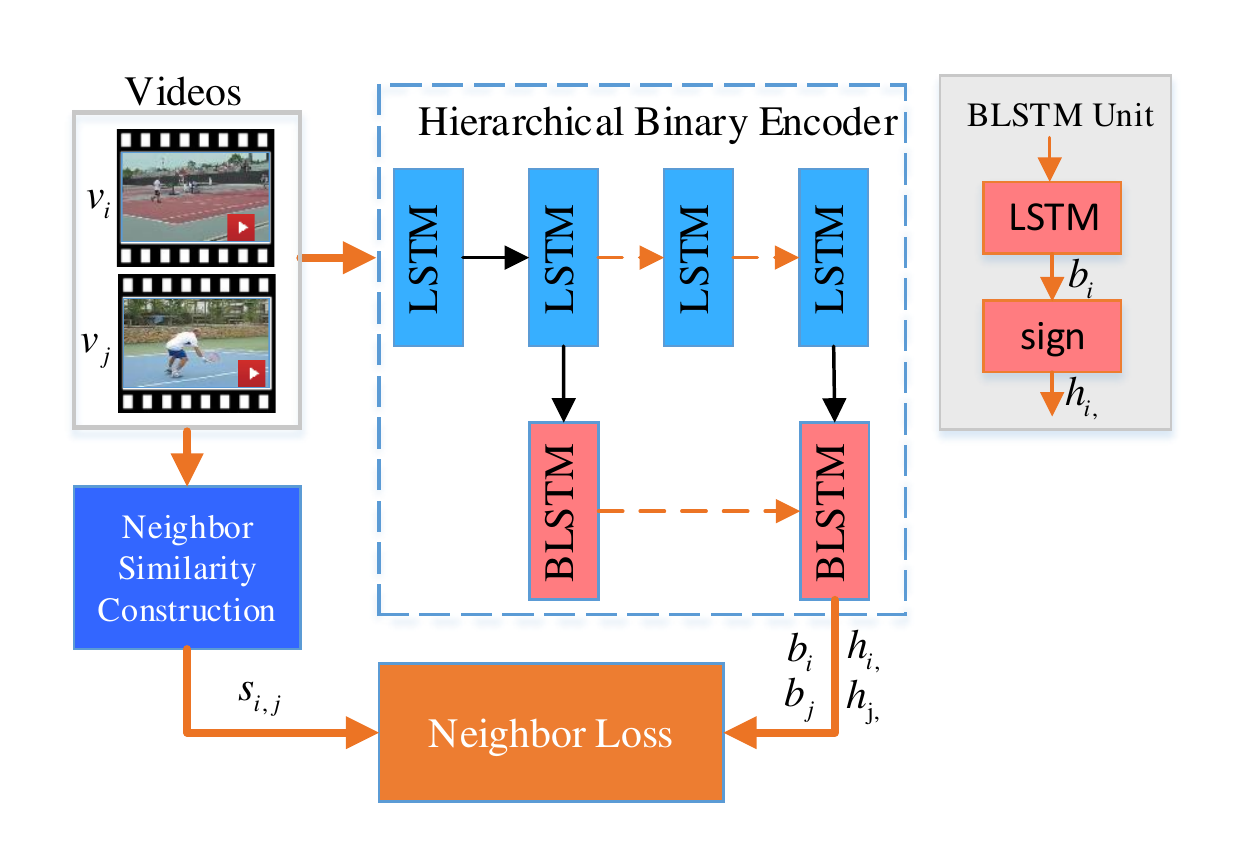}
		\centering
		\caption{{The overview of the neighborhood loss. We first construct the neighborhood structure of the videos. Then, the videos are sent to hierarchical LSTM and generate corresponding binary codes. Finally, the neighborhood loss is calculated based on the neighborhood structure and binary codes.}}
		\label{fig.pairwise}
	\end{figure}
	
	We construct this neighborhood structure as a preprocessing step.\footnote{While we can also construct the neighborhood structure directly using the features during the learning of our neural network, without this preprocessing step of feature extraction, we found that the construction of neighborhood structure is time-consuming, and the updating of neighborhood structures based on the updating of video features in each epoch does not have significant improvement on the performance. Therefore, we fix this neighborhood structure calculated based on the features.}
	First, we use VGG~\cite{simonyan2014very} network to extract frame-level features of videos $ \{\mathbf{v}_{i,k}\}_{k=1}^M$, and use mean-pooling to get a video-level representation ${\mathbf{V_i}} = \frac{1}{M}\sum\nolimits_{k = 1}^M {{\mathbf{v}_{i,k}}}$. We expect that neighboring videos have similar video-level representations. We use cosine similarity to measure the similarity between any two videos $i$ and $j$:
	\begin{equation}
	{CosSim_{i,j}} = \frac{{\mathbf{V_i}^{\rm{T}} \cdot {\mathbf{\mathbf{V_j}}}}}{{\left\| {{\mathbf{V_i}}} \right\|^2\left\| {{\mathbf{V_j}}} \right\|^2}}
	\end{equation}
	For each video, we find its $K_1$-NN and store their indexes in $P\in {{R}^{N \times K_1}}$.
	If we set $K_1$ to a small number, there are not enough neighboring information to preserve for our large dataset. On the other hand, if we simply increase $K_1$, the accuracy of the retrieved neighbors drops.
	Therefore, we design a new strategy to obtain more neighborhood structure information. 
	Specifically, we compute the intersection of a video's most relevant videos' indexes then we get the top $K_2$ indexes based on the size of intersection. Every video in the $K_2$ indexes shares at least one common neighbor with the video. For example, after retrieving $K_1$-NN, we get the relevant video indexes of video $i$ as $\{1,2,3,4,5\}$ and the relevant video indexes of video $j$ as $\{1,3,5,7,9\}$. If the size of intersection is among the top $K_2$-NN of video $i$, all videos in the intersection will be neighbors of video $i$, which means $\{1,2,3,4,5,7,9\}$ will be regarded as neighbors of video $i$. We construct the similarity matrix $S\in {R}^{N\times N}$ by preprocessing the train data as mentioned above. If $s_{i,j} = 1$, video $i$ and video $j$ are considered as neighbors. And $s_{i,j} = -1$ means video $i$ and video $j$ are not neighbors.

	\textbf{Neighborhood Structure Loss}:
	As described in Sec.\ref{subsec.hbae}, we can get the binary codes $ \mathbf{B} = \{\mathbf{b}_i\}^L$ from binary encoder for all the videos. Then we define a loss function based on the neighborhood structure as:
	\begin{equation}
	\min L = 
	{\sum\limits_{{s_{i,j}} \in S} {(\frac{1}{L}\mathbf{b}_i^T{\mathbf{b}_j} - {s_{i,j}})} ^2}
	\label{equ.pairwise}
	\end{equation}
	where $b_{i},b_j$ are the hash codes of videos $i$ and $j$, $s_{i,j}$ indicates the similarity of video $i$ and $j$. $\mathbf{b}_i = sgn(\mathbf{h}_i)$, and $\mathbf{h}_i$ is the hidden state of encoder before binarization of video $i$ as in Eq.\ref{equ.sgn}. Instead of defining the loss function on the binary codes $\mathbf{b}_i$ and $\mathbf{b}_j$, we put the constraints of neighborhood structure preserving on the $\mathbf{h}_i$ and $\mathbf{h}_j$. Then, we have the following regularized problem by replacing the equality constraint in Eq.\ref{equ.pairwise} by a regularization terms as:
	\begin{equation}
	\begin{aligned}
	NeighborLoss_{i,j} = 
	{\sum\limits_{{s_{i,j}} \in S} {(\frac{1}{L}\mathbf{h}_i^T{\mathbf{h}_j} - {s_{i,j}})} ^2} + 
	\eta {(||{\mathbf{b}_i} - {\mathbf{h}_i}||)^2 } 
	\label{equ.pairwise_reg}
	\end{aligned}
	\end{equation}
	where $\eta$ is weight of regularization term. This term impels neighboring videos to have similar hash codes. The overview is illustrated in Fig.\ref{fig.pairwise}.
	The final loss function is composed of the reconstruction loss as Eq.\ref{equ.recon} and pairwise loss as Eq.\ref{equ.pairwise_reg}:
	\begin{equation}
	Loss \!=\! \lambda \!\times\! \sum\limits_{i \!=\! 1}^N{ReconLoss_{i}} \!+\! (1-\lambda) \!\times\! \sum\limits_{i,j = 1}^N{NeighborLoss_{i,j}}
	\label{equ.main_loss}
	\end{equation}
	$\lambda$ is a hyper parameter of our model which balances the reconstruction loss and neighbor loss.
	
	\subsection{Optimization Method}
	\label{optim}
	
	In this section, we will formulate our loss function and come up with a scheme to train our model. Suppose that $\theta_e$ denotes the parameters of hierarchical binary encoder. $\theta_d$ denotes the parameters of decoder composed of the forward decoder parameters $\theta_{df}$, backward decoder parameters $\theta_{db}$ and global decoder parameters $\theta_dg$. Binary codes can be obtained by encoder as:
	\begin{equation}	
	\mathbf{b}_i = Encoder(\theta_e, \mathbf{v}_i)
	\end{equation}
	Then $\mathbf{b}_i$ is used to reconstruct video frame features in forward, backward and global mode.
	\begin{align}	
	\bar{\mathbf{v}}_{i} = ForwardDecoder(\theta_{df}, \mathbf{b}_i) \\
	\tilde{\mathbf{v}}_{i} = BackwardDecoder(\theta_{db}, \mathbf{b}_i) \\
	\mathbf{v}_{i,g} = GlobalDecoder(\theta_{dg}, \mathbf{b}_i)
	\end{align}
	As illustrated in Sec.\ref{subsec.hbae}, we can re-write the reconstruction loss as:
	\begin{equation}
	\small
	ReconLoss \!=\! \sum\limits_{t = 1}^M {||{\mathbf{v}_{i,t}}-  \bar{\mathbf{v}}_{i,t}||^2} \!+\! \sum\limits_{t = M}^1 {||{\mathbf{v}_{i,t}}-\tilde{\mathbf{v}}_{i,t}||^2} \!+\! ||{\mathbf{v}_{i,g}}-\tilde{\mathbf{v}}_{i,g}||^2
	\label{equ.recon.detail}
	\end{equation}
	
	\eat{As proved Eq.\ref{equ.pairwise_reg} that parameters of neighbor loss is $\theta_e$.}We can update the parameters $\theta_e$, $\theta_df$, $\theta_{db}$ and $\theta_{dg}$ by utilizing back propagation to optimize our model. 
	
\begin{figure}[t]
		\centering
		\includegraphics[width=0.45\linewidth]{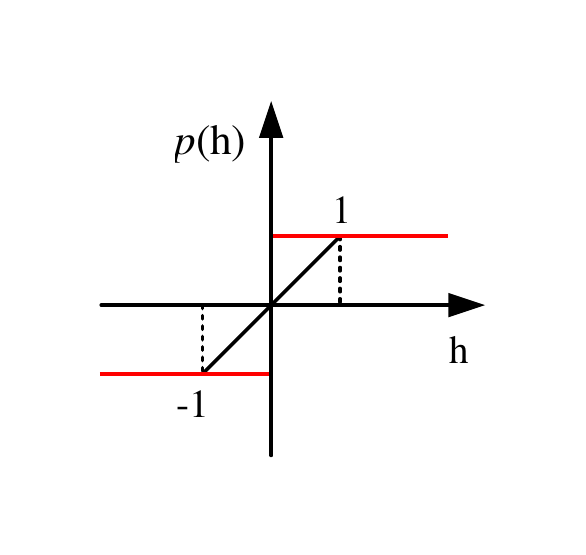}
		\centering
		\caption{{Approximated sgn function using $p(\textbf{h})$. Illustrative process of how $p(\textbf{h})$ (black line) approximates $sgn(\textbf{h})$ (red line).}}
		\label{fig.app_sgn}
	\end{figure}

	However, training SSVH equipped with BLSTM is essential NP-hard as it involves binary optimization of the hash codes that requires combinatorial search space. We follow SSTH~\cite{zhang2016play} to deal with this binary optimization problem and use approximated sgn function as Fig.\ref{fig.app_sgn}:
	\begin{numcases}{sgn(h) \approx p(\mathbf{\mathbf{h}}) =  } \nonumber
	-1, & for $\mathbf{h}<-1$ \\ \nonumber
	\mathbf{h}, & for $-1 \le \mathbf{h} \le 1$\\
	1, & for $\mathbf{h} > 1$
	\label{eq.ph}
	\end{numcases}
	Then we can get the derivative of sgn(h) as :
	\begin{equation}
	{\mathop{\rm sgn}} '(\mathbf{h}): = p'(\mathbf{h}) = 1(|\mathbf{h}| \le 1)
	\end{equation}
	
	The derivative $p'(\mathbf{x})$ states a simple back-propagation rule for BLSTM: when the gradients back propagate to the sgn function, we only allow gradients, whose neural response are between -1 and +1, to pass through.  {Note that we can also utilize other functions, e.g., $tanh(\textbf{h})$ to approximate $sgn(\textbf{h})$. We will evaluate the performance differences in the experiments.}
	
	\subsection{Comparison with SSTH }
	We improve the version of SSTH from \cite{zhang2016play} by defining a hierarchical recurrent structure rather than using a stacked structure. Despite of the improved performance of SSTH, a major disadvantage of stacking is that it introduces a long path from the input to the output video vector representation, thereby resulting in heavier computational cost. Compared with SSTH, SSVH proposed dramatically shortens the path with the capability of adding non-linearity, providing a better trade-off between efficiency and effective. In other words, SSVH extracts the video information at different time scales, and learns the video hash codes with multiple granularities. Moreover, in \cite{zhang2016play}, the encoder RNN with BLSTM runs through the sequence, generating a set of hash codes and then the decoder RNN decodes them to reconstruct the frame-level feature sequence in both forward and reverse orders. Compared with SSTH, we reconstruct the video by not only with forward and backward hierarchical decoders but also with a new global hierarchical binary decoder. Global reconstruction ensure the accuracy of reconstruct appearance information. In addition,  we propose a neighbor structure to further improve the performance. It makes similar videos have similar hash codes and different videos have different hash codes.

	\section{Experiments}
	
	All our experiments for SSVH are conducted with Theano~\cite{jamestheano} on a NVIDIA TITAN X GPU server. Our model can be trained at the speed of about 32 videos per second with a single TITAN X GPU.
	
	\subsection{Datasets and Setting}
	We choose the popular FCVID \cite{jiang2015exploiting} and YFCC \cite{thomee2015new} to evaluate the performance of our model. 
	
	\textbf{FCVID}: FCVID is a large video dataset named Fudan-Columbia Video Dataset containing 91,223 web videos annotated manually according to 239 categories. The categories in FCVID cover a wide range of topics like social events (e.g. ``tailgate party''), procedural events (e.g. ``making cake''), objects (e.g. ``panda''), scenes (e.g. ``beach''), etc. We download 91,185 videos from this dataset because of some damaged videos. This dataset is split into training set containing 45,585 videos and test set containing 45,600 videos. We use the training set for unsupervised learning and the test set as retrieval database and queries.

	\textbf{YFCC}: Yahoo Flickr Creative Commons 100 Million Dataset is a huge collection of multimedia data. There are 0.8M videos in this dataset officially. But we only collected 700,882 videos because of invalid url and corrupted videos. We select 511,044 videos as our dataset from YFCC. We use the 101,256 labeled videos as in \cite{zhang2016play} for retrieval and the left 409,788 unlabeled data as training data.

	\subsection{Implementation Details}
	In this section, we will introduce some implementation details of our model. For each video, we get equally-spaced 24 frames and we think that is enough to represent a video. We use VGG \cite{simonyan2014very} network to extract the frame-level features in our experiment and obtain the 4096-d features as the input of our model. The stride of hierarchical auto-encoder is 2 and the second layer of decoders has 12 units. Due to the huge scale of YFCC, we can not construct a neighbor similarity matrix of 400K$\times$400K. We split the training data into 9 parts because of the limitation of memory and each part have around 45K videos. For each part, we get their neighbor structure matrix. Then we train our model {in order}. Some videos in YFCC (around 50K) do not belong to the 80 categories, and following the instructions of the data provider, we regard these unlabeled videos as the ``others''.
	
	During the training, we use Stochastic Gradient Descent (SGD) algorithm to do parameters updating with a mini-batch size of 256. The regularization parameters are set as $\eta$ = 0.2 in Eq.\ref{equ.pairwise_reg} and $\lambda=0.001$ in Eq.\ref{equ.main_loss}. In neighborhood structure, we choose K1 as 20 and K2 as 10. To compare with baseline methods, we use the publicly available codes and run them on both FCVID and YFCC.

	\subsection{Evaluation Metrics}
	We adopted Average Precision at top K retrieval videos(AP@K) for retrieval performance evaluation \cite{over2014trecvid}. AP means the average of precisions at each correctly retrieved data point. $R$ denotes the number of total relevant videos. $R_i$ means the number of relevant videos, $I_i =1$ means the retrieved video is relevant and $I_i=0$ otherwise. AP@K is defined as $\frac{{\rm{1}}}{{\min (R,K)}}\sum\nolimits_{{\rm{i}} = 1}^K {\frac{{{R_i}}}{i}}  \times {I_i}$. We use the whole test video set as the queries and database. Then we can obtain mAP@K by taking the mean of AP@K of all queries. Hamming ranking is used as the search protocol.
	
	\subsection{Components and Baseline Methods}
	In this subsection, we aim to investigate the effect of each component in our framework. Here we introduce some combinations of the components:
	\begin{itemize}
		\item \textbf{FB} (Forward and Backward Reconstruction): FB consists of forward and backward reconstruction loss $Loss_{f}$ and $Loss_{b}$ as illustrated in Fig.\ref{fig.framework}. FB reconstructs the frame-level features and trains the hash functions simultaneously.
		\item \textbf{FB + GR} (FB + Global Reconstruction (GR)): Based on forward and backward reconstruction loss, we add the global reconstruction loss $Loss_{g}$.
		\item \textbf{Neighborhood Structure.} Neighborhood structure loss $NeighborLoss$ preserves the neighborhood structure in the original space, as is illustrated in Sec.\ref{sub.sec.pairwiseModel}.
		\item \textbf{SSVH.} SSVH is composed of the $Loss_{f}$, $Loss_{b}$, $Loss_{g}$ and $NeighborLoss$, which is illustrated in Fig.~\ref{fig.framework}.
		\item \textbf{FB + GR + GTHNS.}(FB + GR + Groundtruth Neighborhood Structure) Here, we compute the neighbor similarity using the labels of training data to compare with our SSVH.	
	\end{itemize}
	We also compare our method with several the state-of-the-art unsupervised hashing methods to validate the performance of our method. These methods are:
	\\\noindent \textbf{ITQ.} Iterative Quantization (ITQ) \cite{DBLP:journals/pami/GongLGP13} is a representative unsupervised hashing method for image retrieval, and we extend it for video retrieval. We get a video-level feature by applying mean-pooling on the frame-level features. 
	\\\noindent \textbf{Submod.} Submodular Video Hashing (Submod) \cite{cao2012submodular} is also a common video retrieval method. We first use mean pooling to extract video representation then hash it into a 1024-dimension code using traditional hashing method LSH. Then we measure the informativeness of training data to select $k$ most informative hash functions.
	\\\noindent \textbf{MFH.} Multiple feature hashing (MFH) \cite{song2011multiple} learns hash functions based on the similarity graph of the frames. It learns frame-level hash codes and uses average pooling to get the real-valued video-level representation, followed by binarization to get the hash codes.
	\\\noindent \textbf{DH.} Deep hashing (DH) \cite{erin2015deep} learns hash functions based on a deep neural network by adding a binarization loss. We use original encoder-decoder model to extract video representations by getting the output of the encoder.
	\\\noindent \textbf{SSTH.} Self-Supervised Temporal Hashing (SSTH) \cite{zhang2016play} means Self-Supervised Temporal Hashing, which trains hash functions as an auto-encoder to reconstruct the frame features in forward and backward order.

	\begin{table*}[t]
		\centering
		\caption{Effect of Components of SSVH on the FCVID dataset (code length=256).}
		\label{tab.components}
		\begin{tabular}{c|c|c|c|c|c}
			\hline
			mAP@K & 20 & 40 & 60 & 80 & 100 \\ \hline
			SSTH & 28.37\% & 24.12\% & 21.26\% & 19.51\% & 17.93\% \\ \hline
			FB(Forward+Backward) & 30.90\% & 24.52\% & 21.25\% & 19.05\% & 17.38\% \\ \hline
			FB + GR(Global Reconstruction) & 31.72\% & 25.36\% & 22.09\% & 19.87\% & 19.18\% \\ \hline
			Neighborhood Structure & 36.51\% & 32.40\% & 30.15\% & 28.40\% & 26.82\% \\ \hline
			\textbf{FB + GR + Neighborhood Structure} & \textbf{37.92\%} & \textbf{33.40\%} & \textbf{30.92\%} & \textbf{29.00\%} & \textbf{27.29\%} \\ \hline \hline
			FB + GR + GroundTruth & 42.41\% & 39.27\% & 37.67\% & 36.38\% & 35.17\% \\ \hline
		\end{tabular}
	\end{table*}
	
	\begin{figure}[t]
		\centering
		\includegraphics[width=0.9\linewidth]{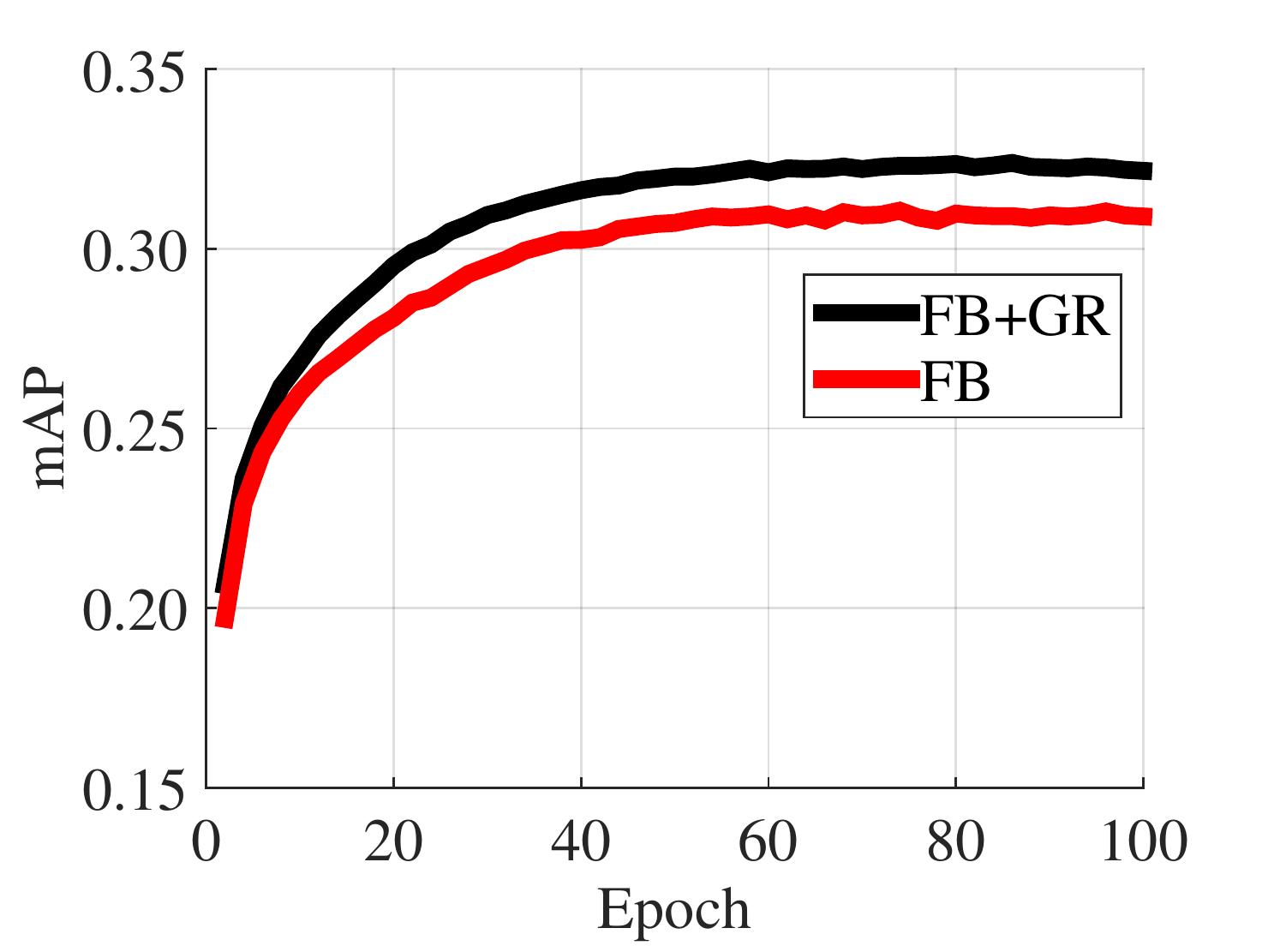}
		\centering
		\caption{The convergence study of our method on the FCVID dataset. FB: Forward+Backward reconstruction, FB + GR: FB + Global Reconstruction. (code length=256)}
		\label{fig.generalloss}
	\end{figure}

		\begin{table*}[t]
		\centering
		\caption{MAP results (i.e., top 5, 10, 20, 40, 60, 80 and 100 retrieval results) on the FCVID dataset of 256bits. The performance variance with different parameters: $K1$ or $K2$.}
		\label{tab.K1_K2}
		\begin{tabular}{l|c|c|c|c|c|c|c|c}
			\hline
			\multirow{7}{*}{K2=10} & K1 & 5 & 10 & 20 & 40 & 60 & 80 & 100 \\ \cline{2-9} 
			& 5\_10 & 52.76\% & 42.46\% & 35.21\% & 29.66\% & 26.76\% & 24.70\% & 23.05\% \\ \cline{2-9} 
			& \textbf{10\_10} & \textbf{54.75\%} & \textbf{45.53\%} & \textbf{39.25\%} & \textbf{34.39\%} & \textbf{31.71\%} & \textbf{29.66\%} & \textbf{27.91\%} \\ \cline{2-9} 
			& 20\_10 & 53.10\% & 44.09\% & 38.10\% & 33.63\% & 31.19\% & 29.29\% & 27.58\% \\ \cline{2-9} 
			& 30\_10 & 52.08\% & 43.06\% & 37.06\% & 32.56\% & 30.12\% & 28.21\% & 26.57\% \\ \cline{2-9} 
			& 40\_10 & 51.31\% & 42.19\% & 36.13\% & 31.72\% & 29.40\% & 27.58\% & 25.99\% \\ \cline{2-9} 
			& 50\_10 & 50.19\% & 40.37\% & 35.78\% & 31.03\& & 28.63\% & 26.83\% & 24.76\% \\ \hline
			\multirow{7}{*}{K1=10} & K2 & 5 & 10 & 20 & 40 & 60 & 80 & 100 \\ \cline{2-9} 
			& 10\_5 & 53.03\% & 42.82\% & 35.67\% & 30.09\% & 27.16\% & 25.06\% & 23.39\% \\ \cline{2-9} 
			& \textbf{10\_10} & \textbf{54.75\%} & \textbf{45.53\%} & \textbf{39.25\%} & \textbf{34.39\%} & \textbf{31.71\%} & \textbf{29.66\%} & \textbf{27.91\%} \\ \cline{2-9} 
			& 10\_20 & 53.18\% & 44.19\% & 38.25\% & 33.74\% & 31.26\% & 29.32\% & 27.60\% \\ \cline{2-9} 
			& 10\_30 & 52.13\% & 43.00\% & 36.99\% & 32.56\% & 30.15\% & 28.29\% & 26.69\% \\ \cline{2-9} 
			& 10\_40 & 51.47\% & 42.26\% & 36.25\% & 31.85\% & 29.52\% & 27.73\% & 26.16\% \\ \cline{2-9} 
			& 10\_50 & 51.18\% & 41.18\% & 35.72\% & 31.22\% & 28.85\% & 27.07\% & 25.57\% \\ \hline
		\end{tabular}
	\end{table*}

	\subsection{Performance Analysis}
	\subsubsection{Effect of Components of SSVH}
	We first test different combinations of the components of our SSVH on the FCVID dataset. While it is impossible to test all the combinations due to the space limit, we focus on the following aspects: 1) Is the hierarchical structure better than stacked LSTM? 2) What is the effect of each component? and 3) What is the performance of our neighborhood structure compared with human labels?

	To achieve this, we report the mAP@K ($K=20,40,60$, $80,100$) results of different combinations in Table.\ref{tab.components}.  We can observe that:
	1) Compared to SSTH, a stacked RNN structure which also uses forward and backward reconstructions, our hierarchical structure (FB) obtains superior performance with less computational cost. 
	2) Using FB only, our method can achieve promising results, and global loss can further improve the performance. By adding GR to FB, the performance is improved by about 1\% for the mAP at different K. GR can also makes the training more stable, as shown in Fig. \ref{fig.generalloss}.
	Neighbor model also obtains excellent performance which proves that it is beneficial to exploit the neighborhood structure of the training data for the task of video retrieval. SSVH achieves the best performance compared with other combinations, which indicates that each component contributes to the good performance of SSVH. 
	3) Besides, we also construct the similarity matrix using the labels. Obviously, our model with human labels significantly outperforms our SSVH with the unsupervised neighborhood structure, which means that we can improve the accuracy of predicted neighbor similarity to enhance our performance.
	
	\begin{figure}[h]
		\centering
		\includegraphics[width=0.85\linewidth]{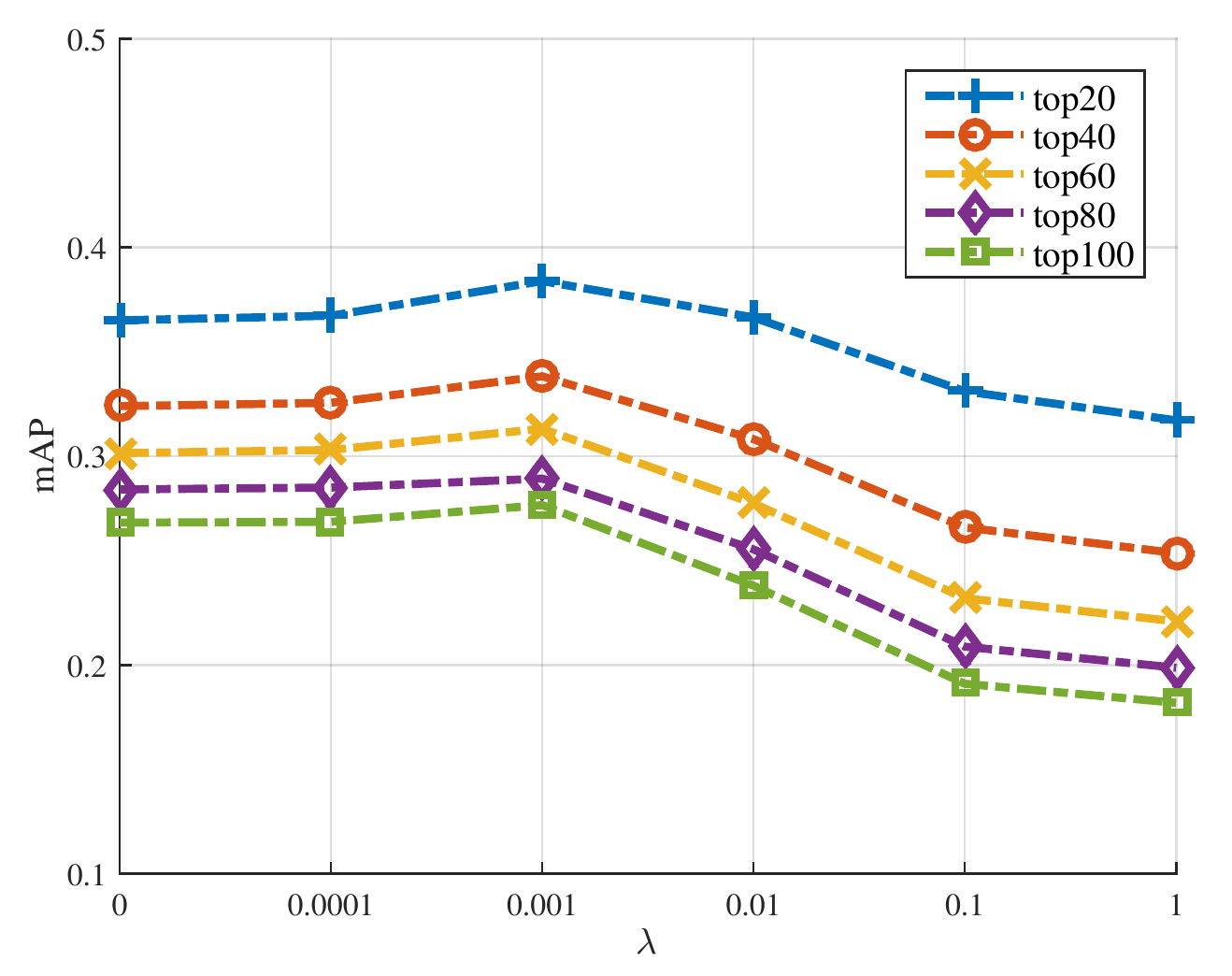}
		\centering
		\caption{{Performances of different $\lambda$ on FCVID (code length=256)}}
		\label{fig.lamb}
	\end{figure}

	\begin{figure*}[t]
		\centering
		\subfigure[FCVID 8 bits]{
			\includegraphics[width=0.30\linewidth]{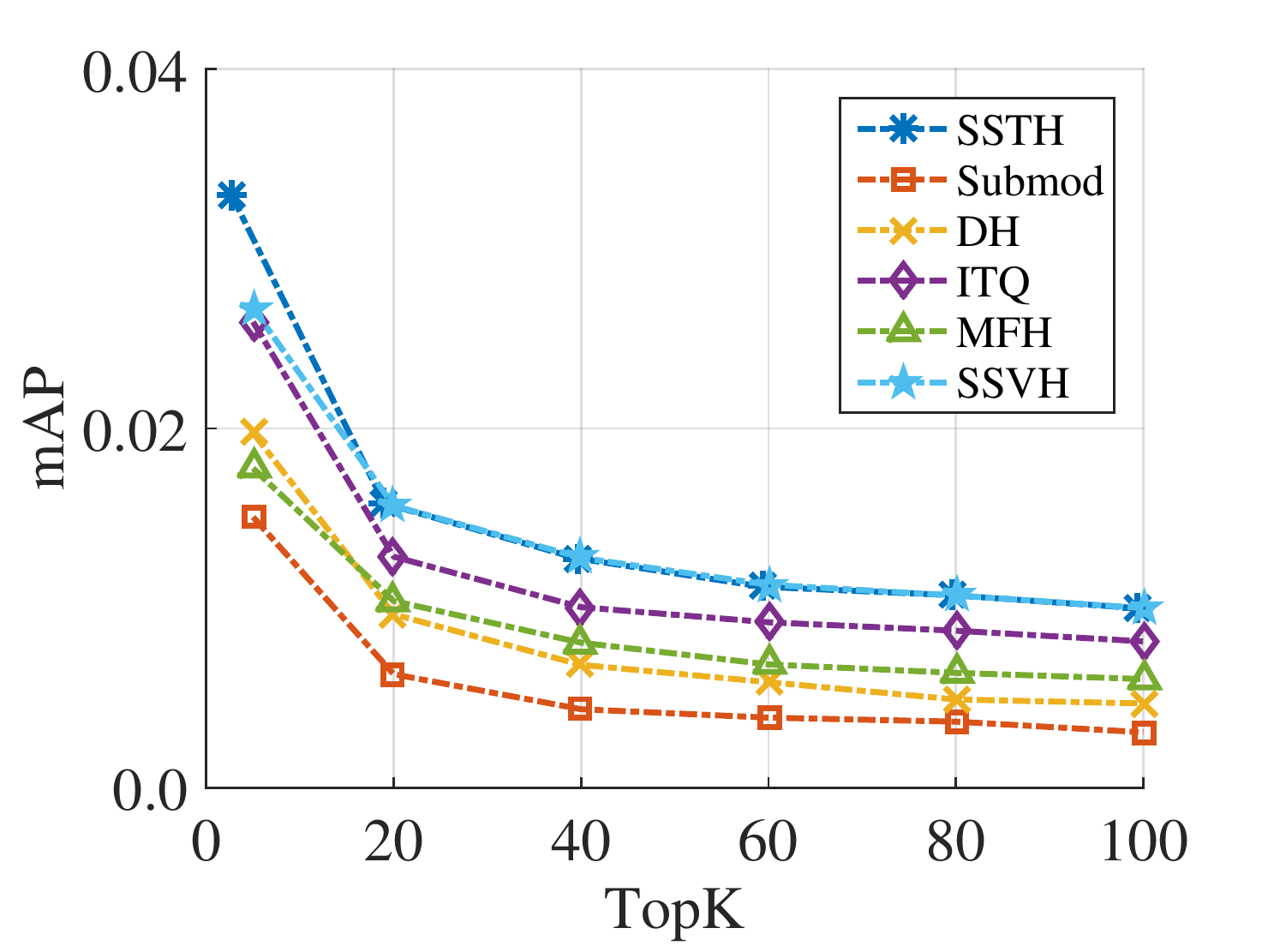}}
		\subfigure[FCVID 16 bits]{
			\includegraphics[width=0.30\linewidth]{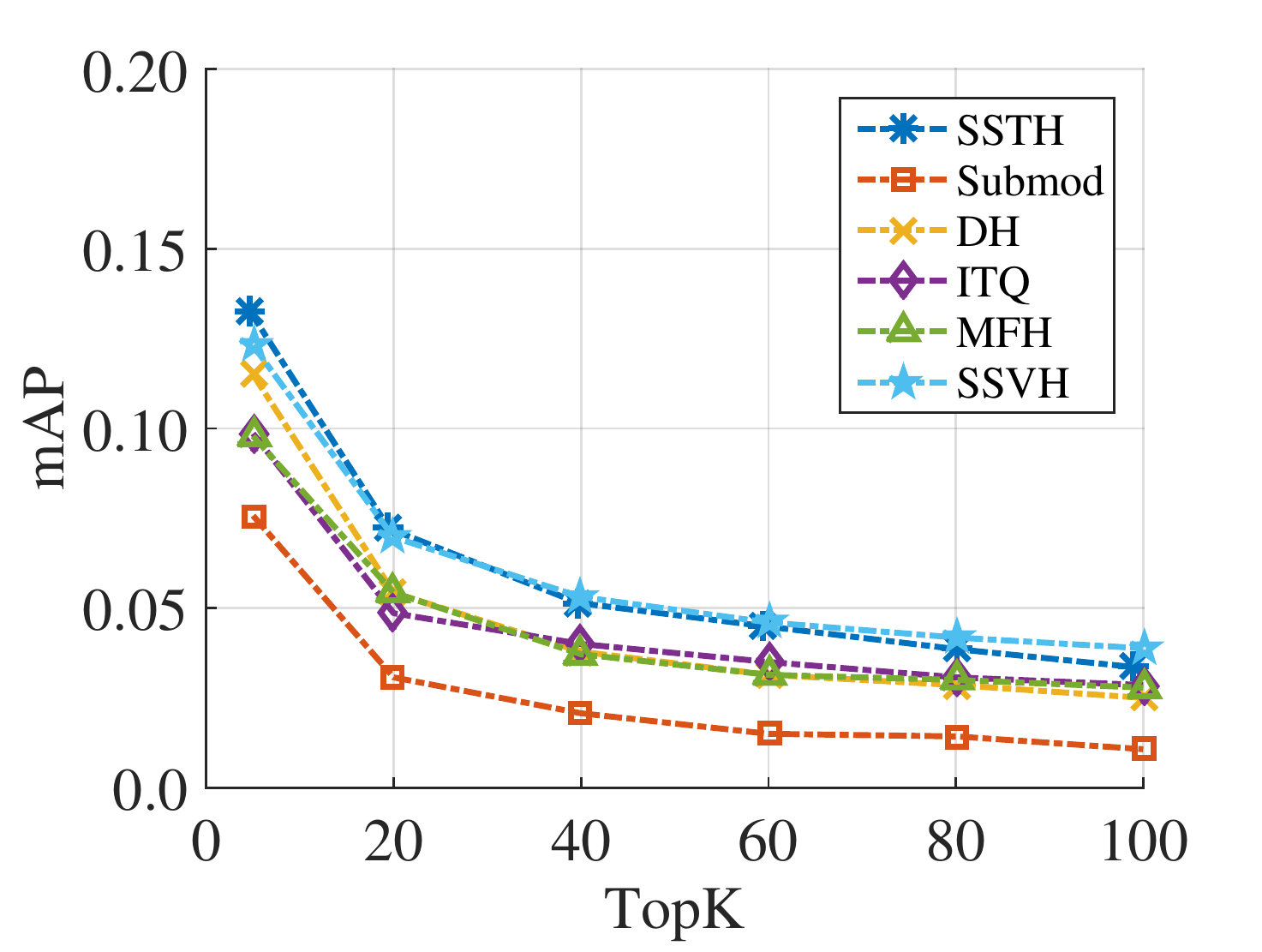}}
		\vspace{-0.17cm}
		\subfigure[FCVID 32 bits]{
			\includegraphics[width=0.30\linewidth]{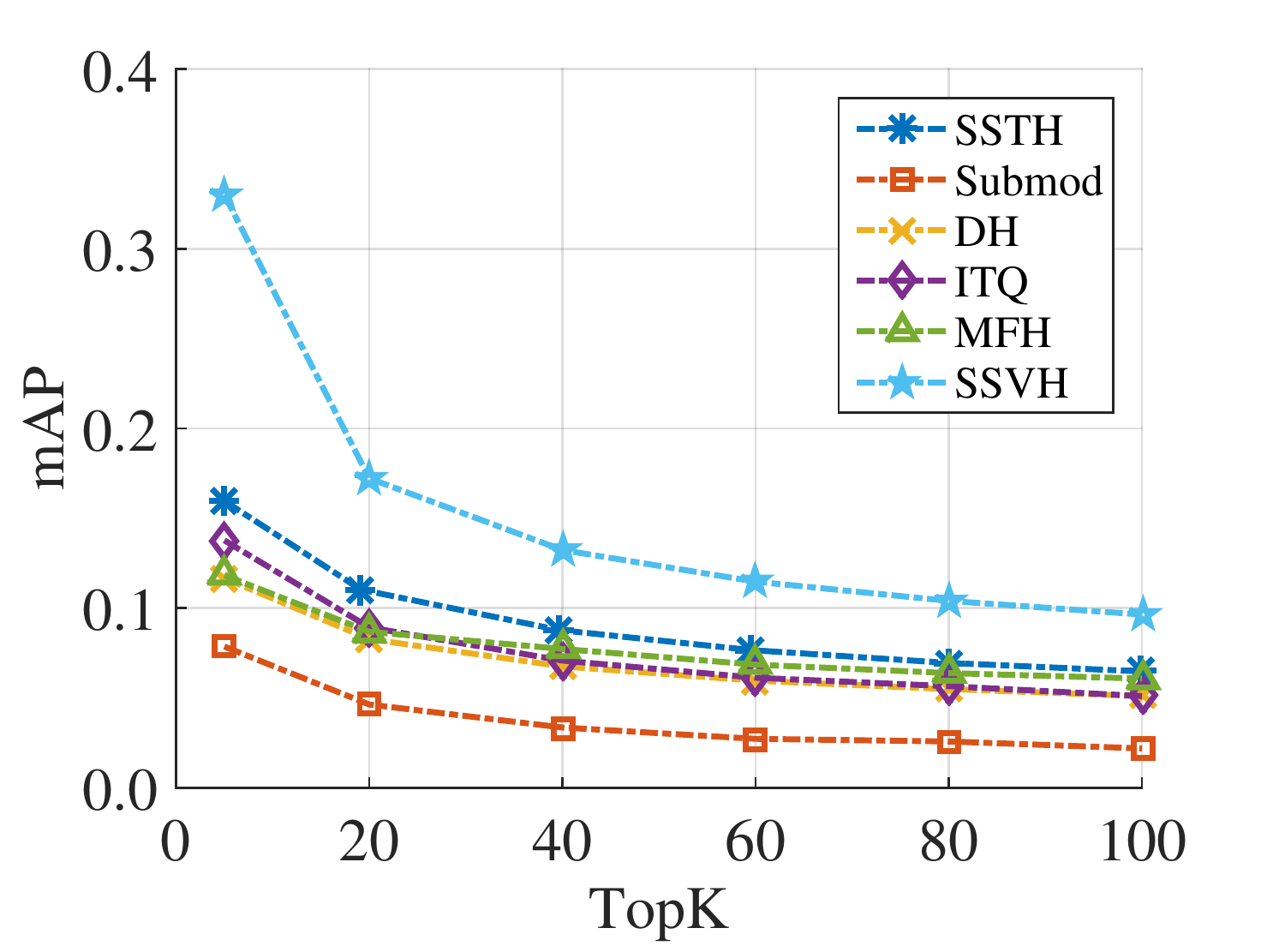}}
		\vspace{-0.17cm}
		
		\subfigure[FCVID 64 bits]{
			\includegraphics[width=0.30\linewidth]{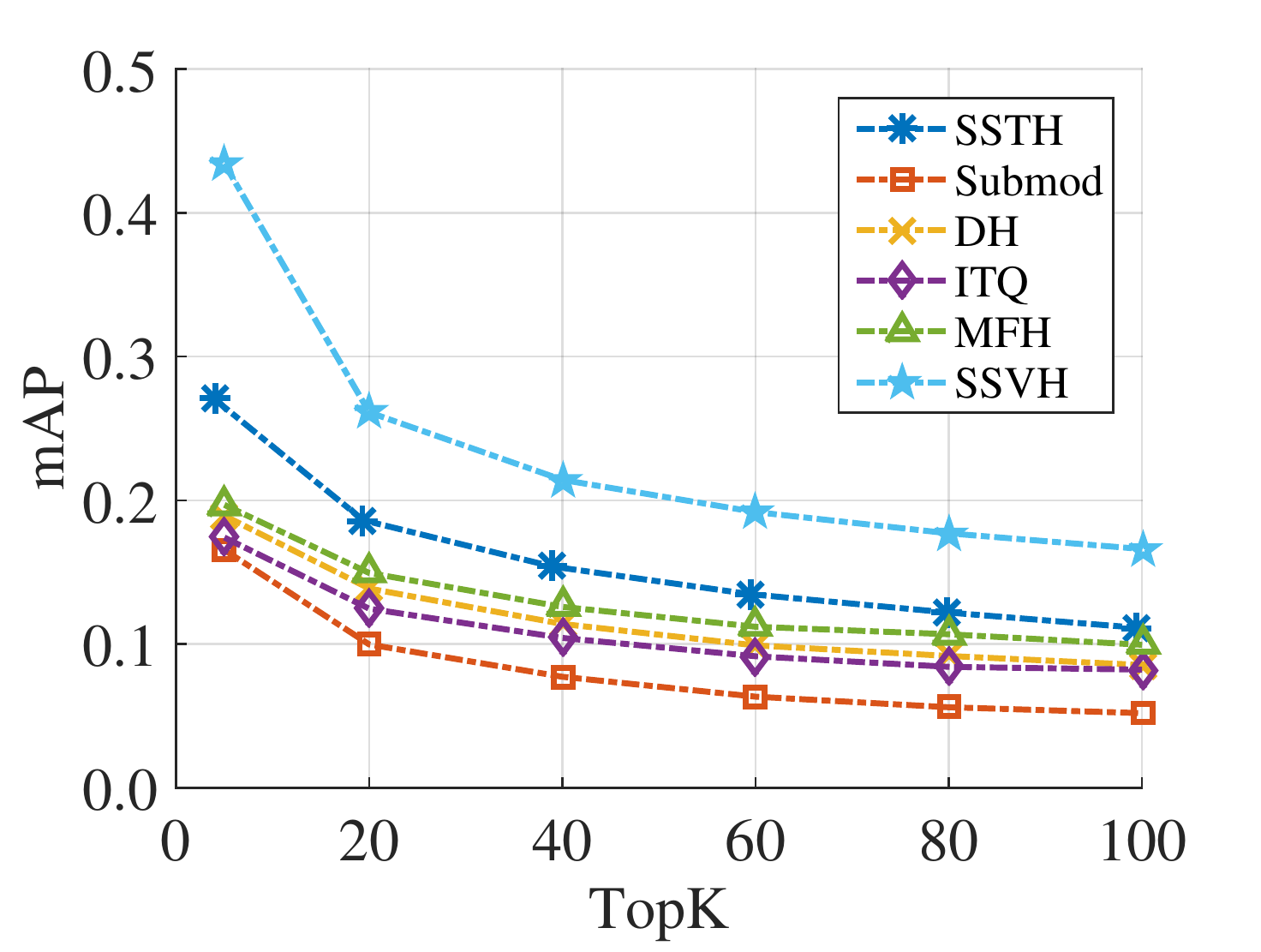}}
		\vspace{-0.17cm}
		\subfigure[FCVID 128 bits]{
			\includegraphics[width=0.30\linewidth]{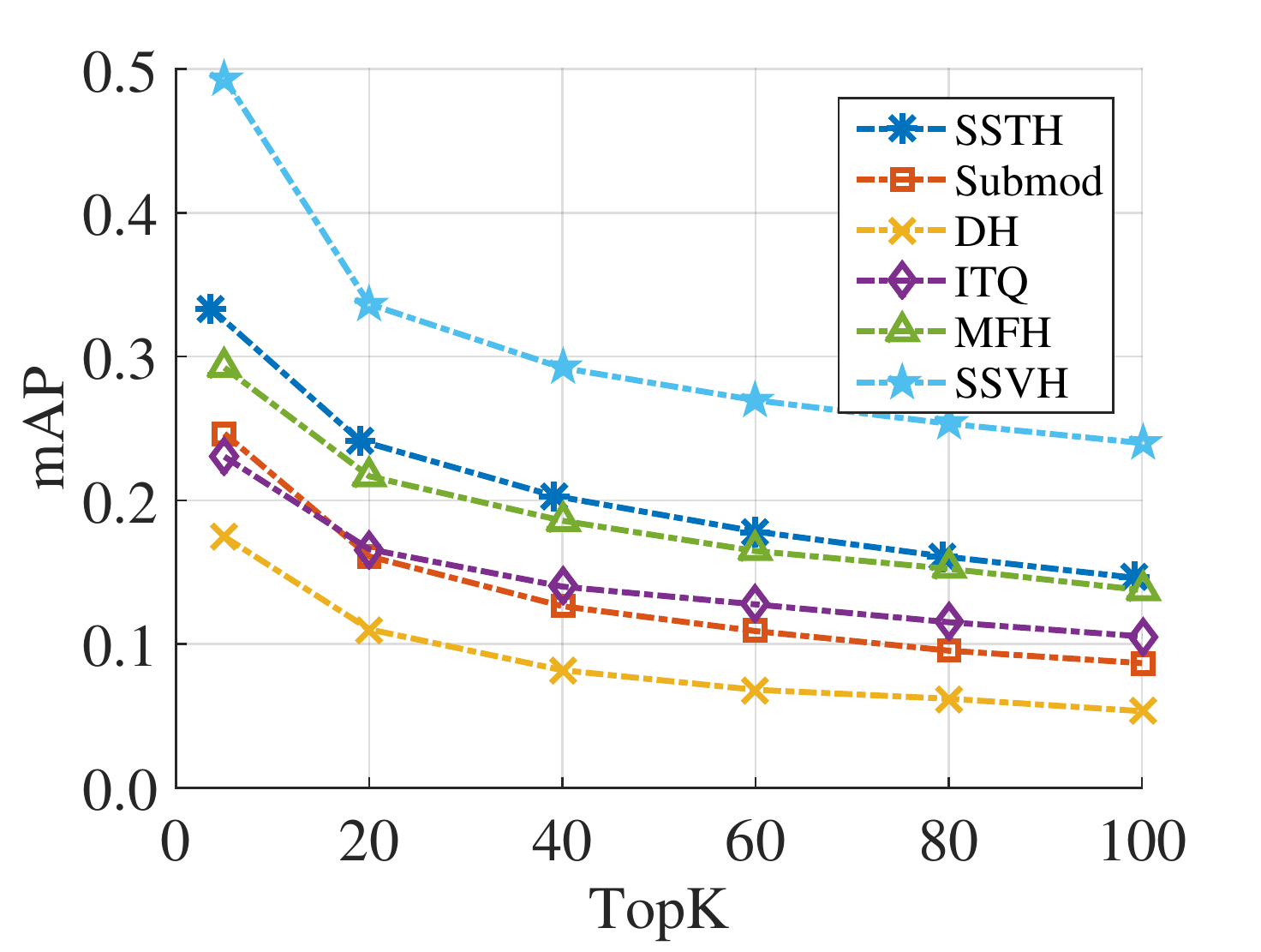}}
		\vspace{-0.17cm}
		\subfigure[FCVID 256 bits]{
			\includegraphics[width=0.30\linewidth]{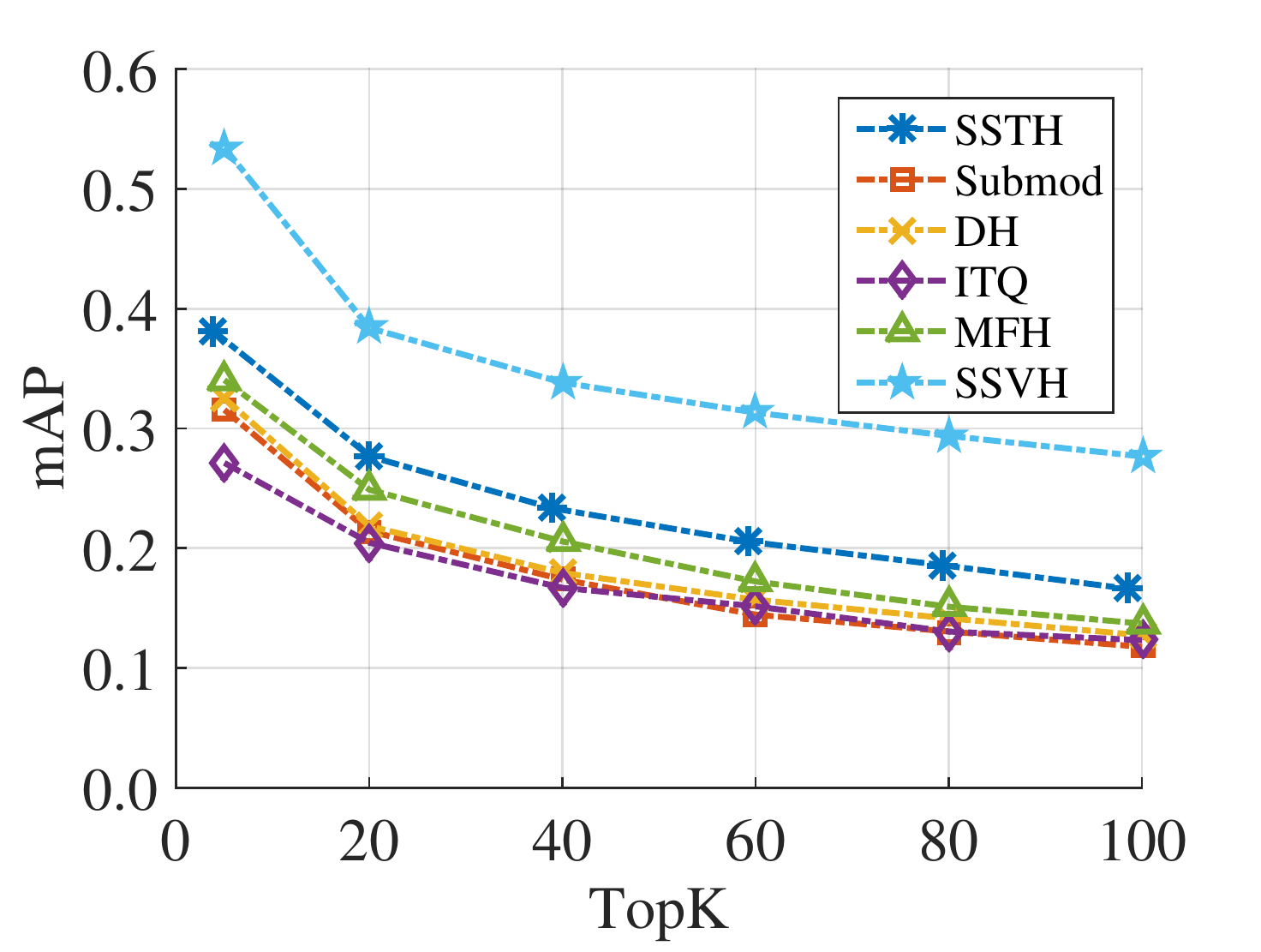}}
		\hspace{-0.17cm}
		
		\subfigure[YFCC 8 bits]{
			\includegraphics[width=0.30\linewidth]{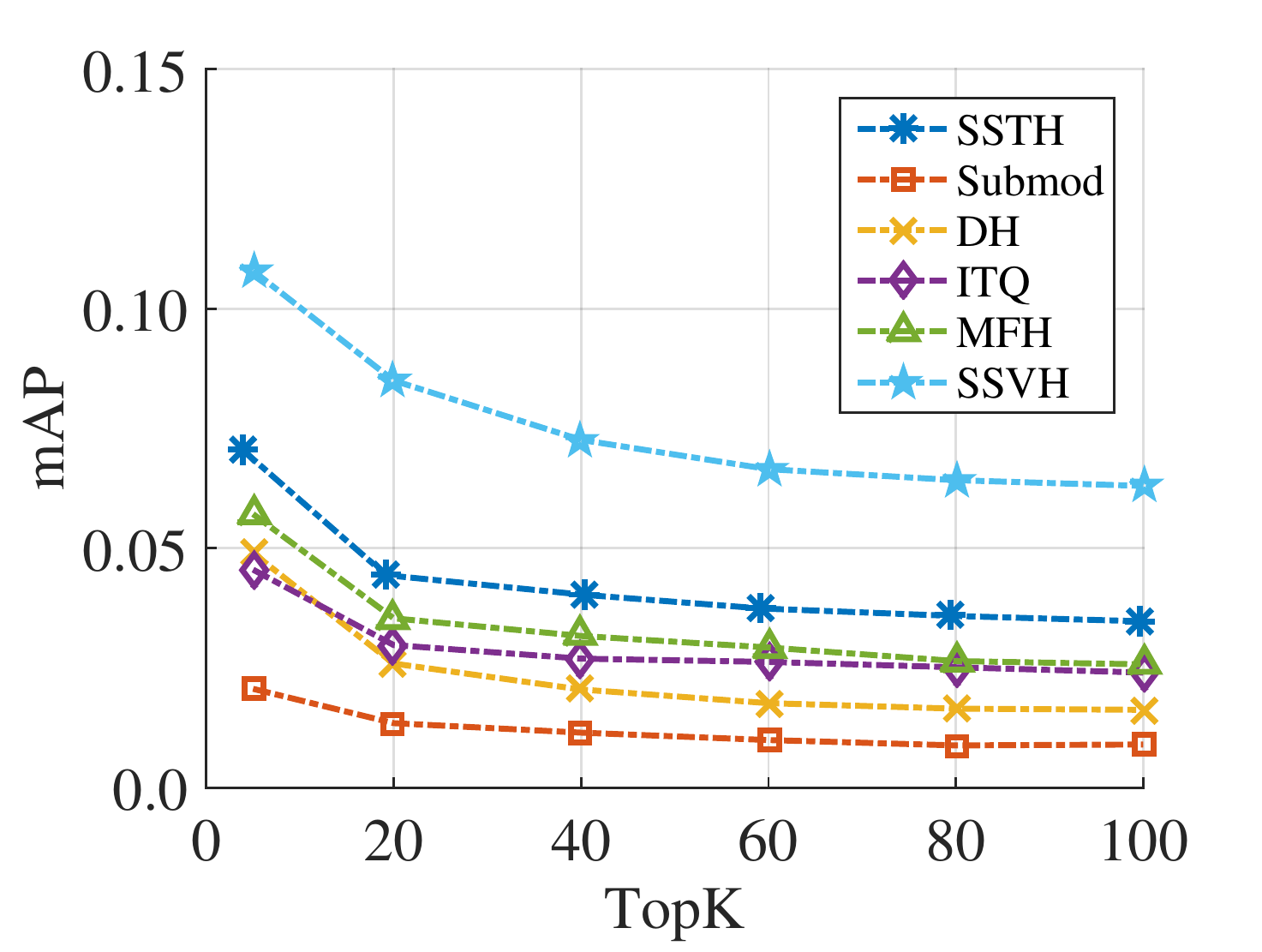}}
		\vspace{-0.17cm}
		\subfigure[YFCC 16 bits]{
			\includegraphics[width=0.30\linewidth]{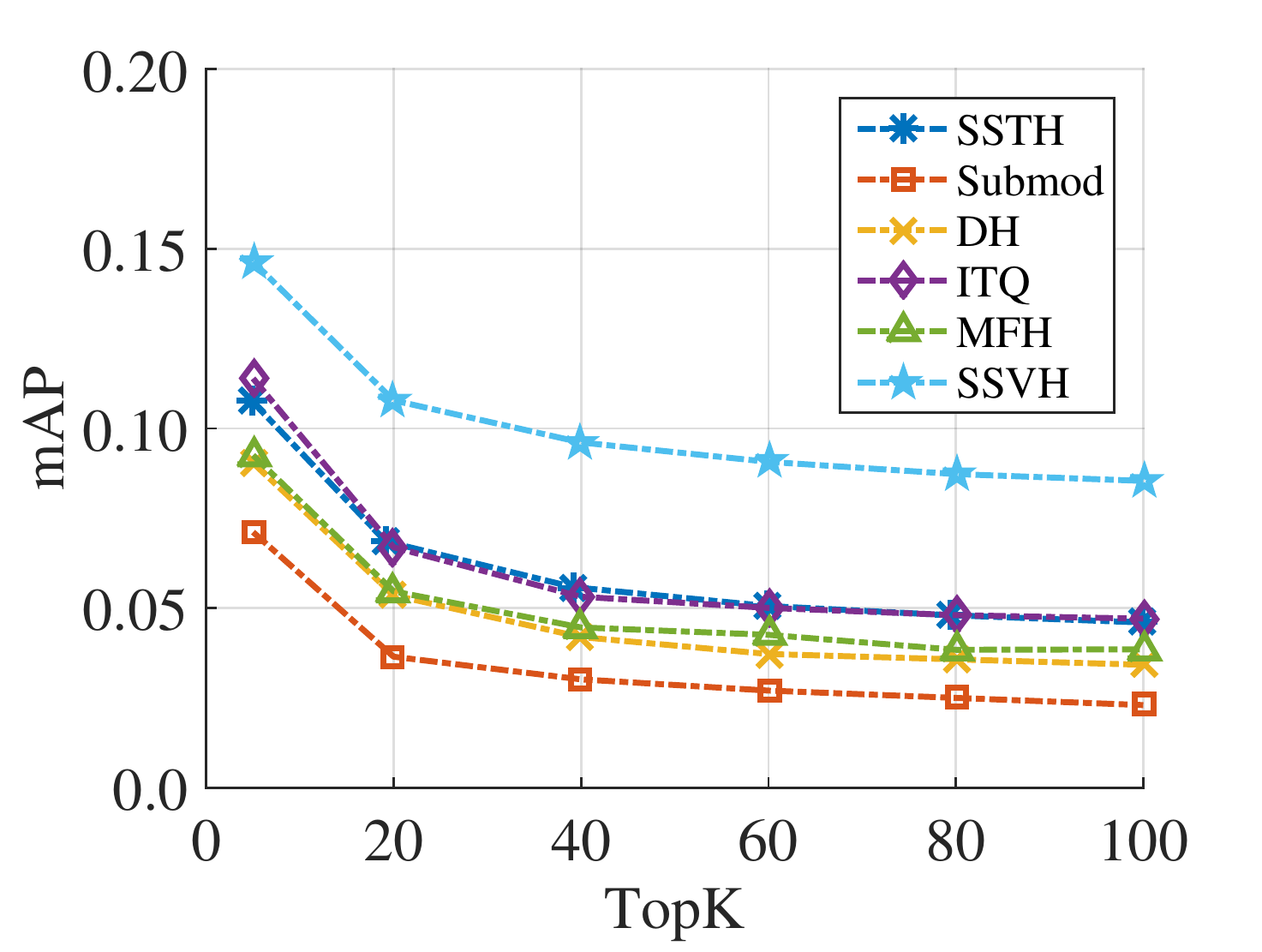}}
		\vspace{-0.17cm}
		\subfigure[YFCC 32 bits]{
			\includegraphics[width=0.30\linewidth]{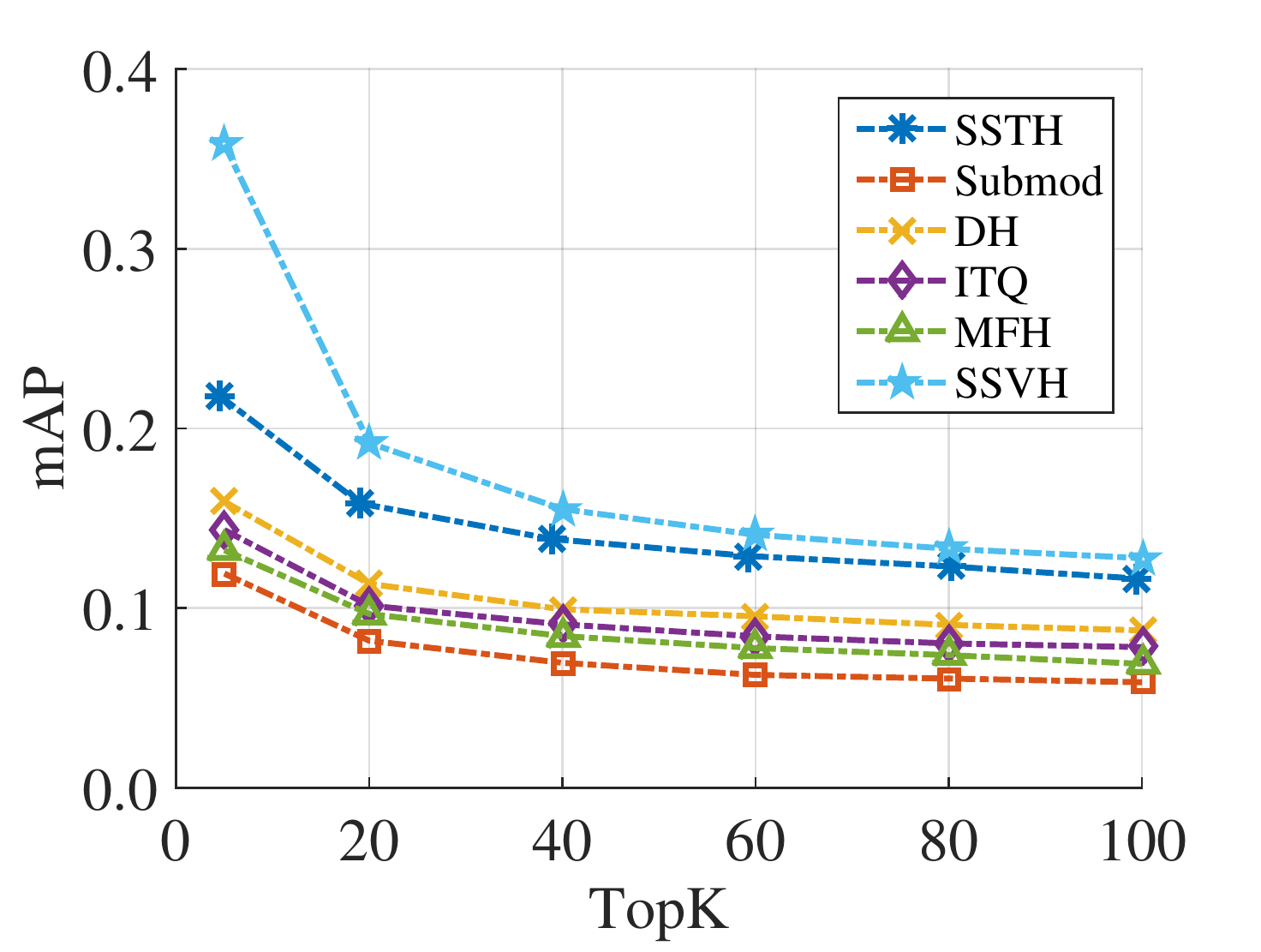}}
		
		\subfigure[YFCC 64 bits]{
			\includegraphics[width=0.30\linewidth]{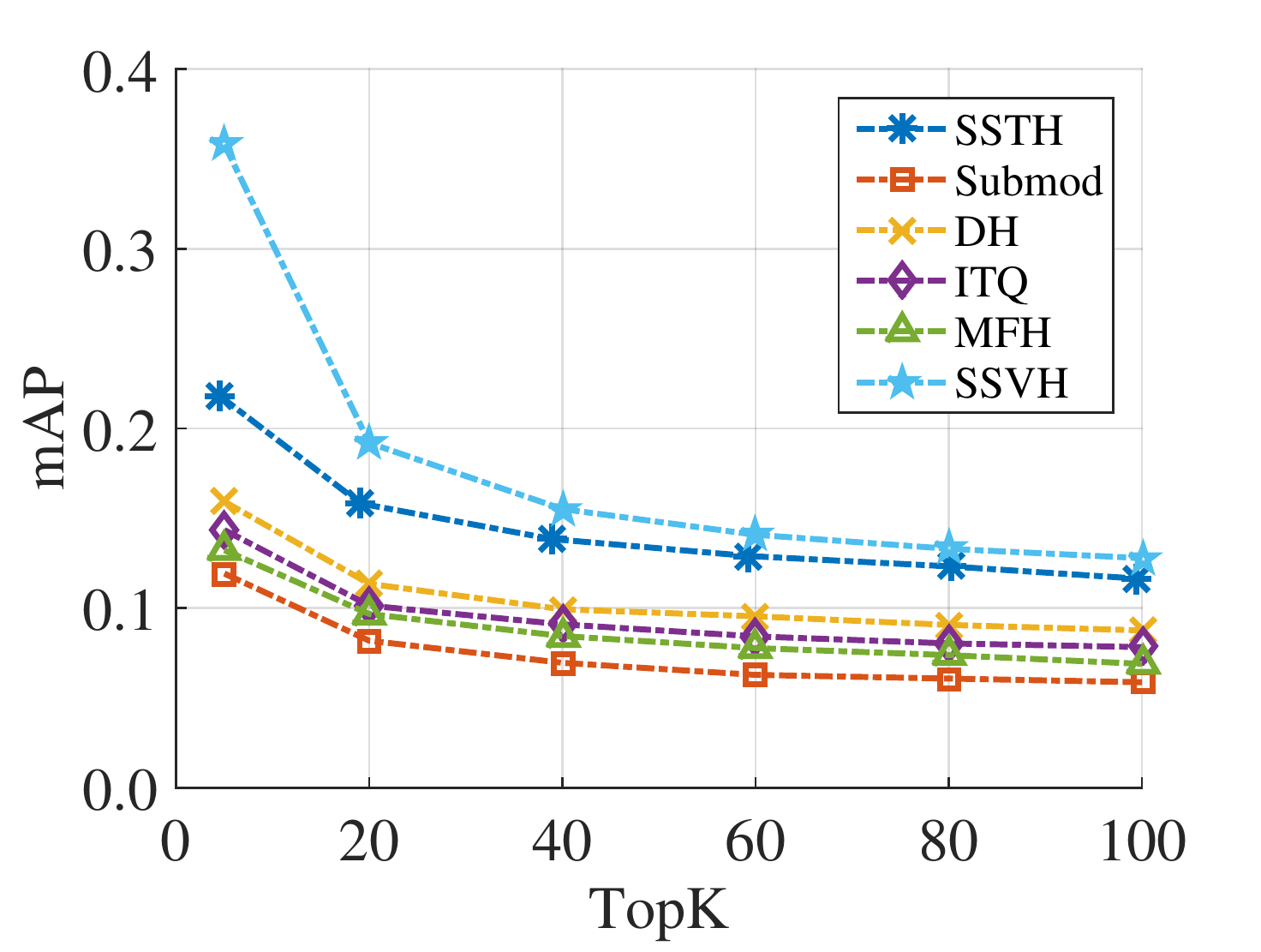}}
		\vspace{-0.17cm}
		\subfigure[YFCC 128 bits]{
			\includegraphics[width=0.30\linewidth]{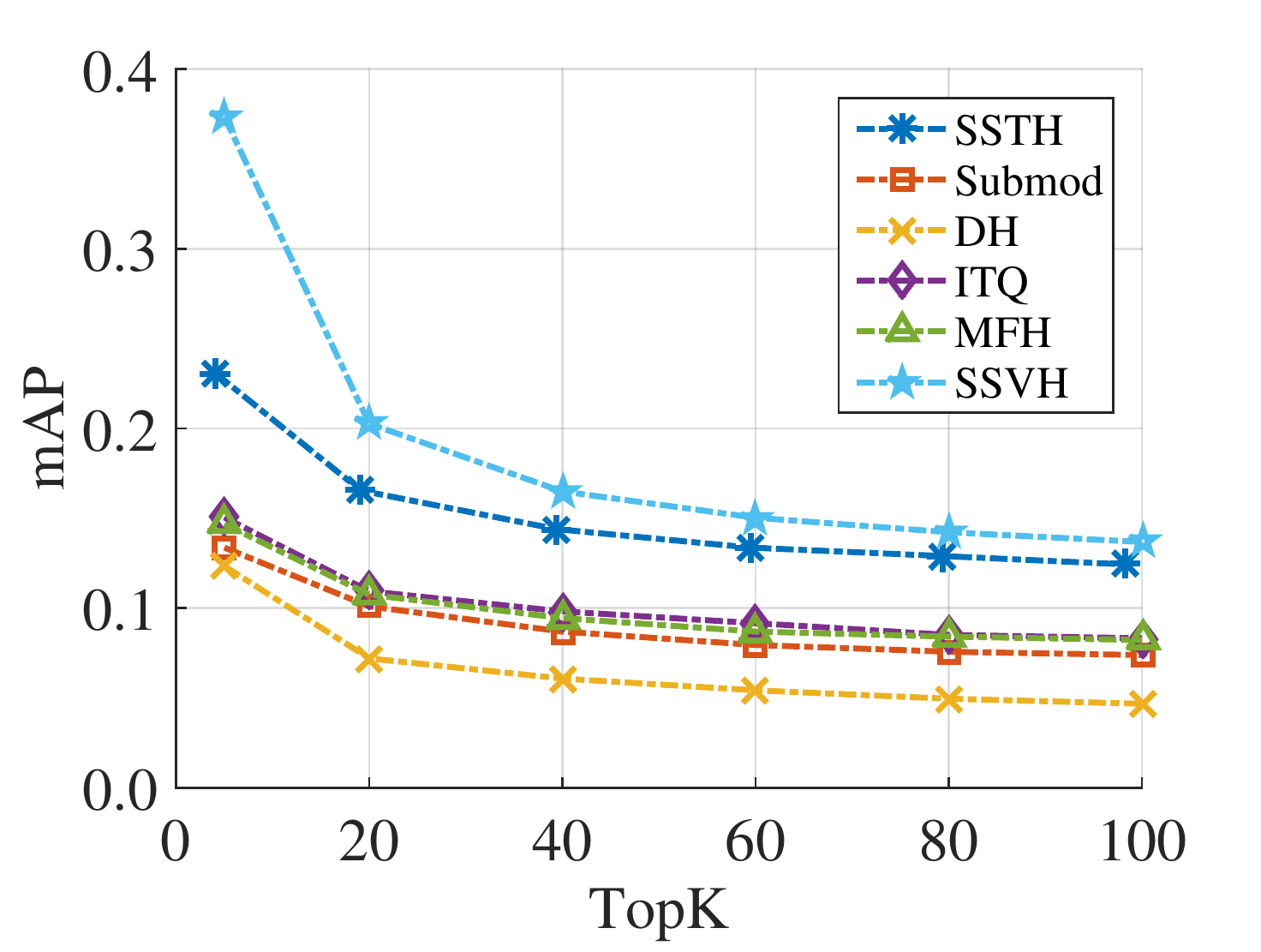}}
		\vspace{-0.17cm}
		\subfigure[YFCC 256 bits]{
			\includegraphics[width=0.30\linewidth]{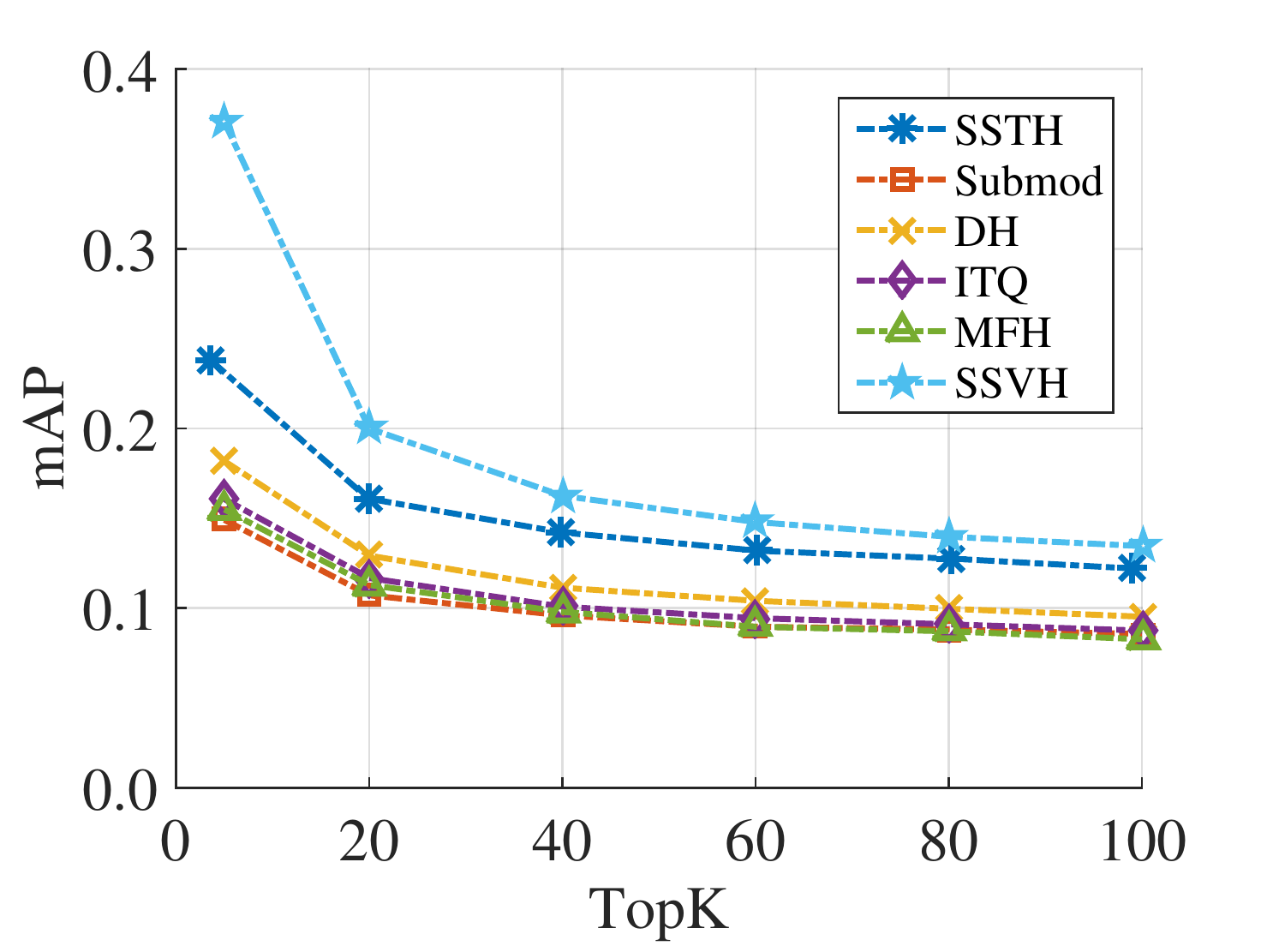}}
		
		\caption{{Performance (mAP@K) of different video hashing methods with different code lengths. The first two rows are the results for the FCVID dataset, and the bottom two rows are for the YFCC dataset.}}
		\label{fig.alg.hash}
	\end{figure*}

	\begin{figure*}[t]
		\centering
		\includegraphics[width=1.0\textwidth]{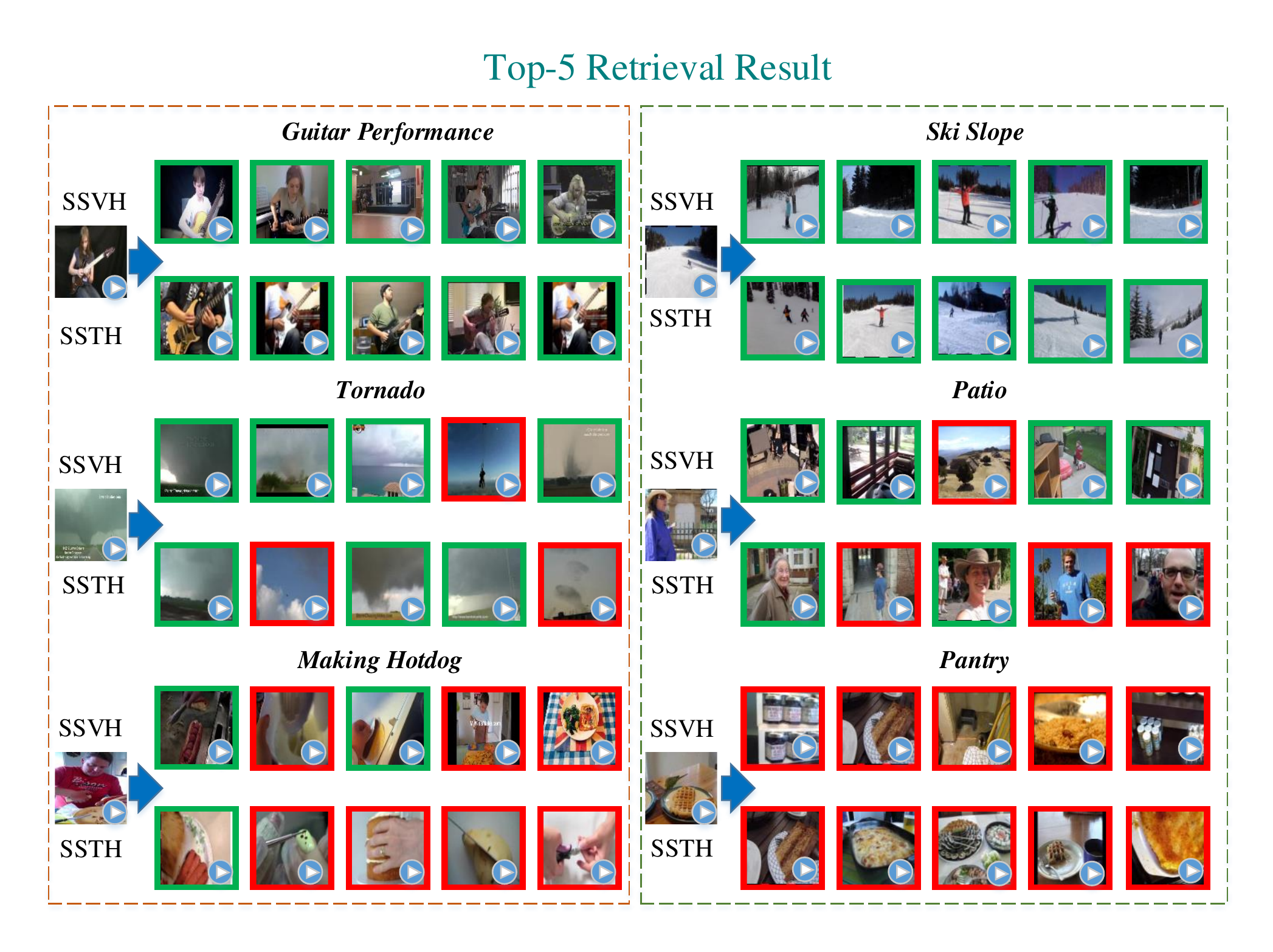}
		\centering
		\caption{The retrieval results of 256 bits when using SSVH and SSTH on FCVID and YFCC. Green border means correct retrieval result and Red border means incorrect retrieval result. Left: FCVID; Right: YFCC}
		\label{fig.performance}
	\end{figure*}

	\subsubsection{Trade off between Neighbor Loss and Reconstruction Loss}
	The hyper-parameter $\lambda$ in Eq.\ref{equ.main_loss} is crucial in our method, which balances the neighborhood similarity loss and the reconstruction loss of the training videos. Therefore, we tune the parameter of $\lambda$ from $0,10^{-4},10^{-3},10^{-2},10^{-1},1$ and show the performance in Fig.\ref{fig.lamb}. The curves in Fig.\ref{fig.lamb} shows how the mAP at top20, top40, top60, top80, and top100 varies with respect to $\lambda$. 
	When $\lambda=0$ and $\lambda=1$, i.e., only the neighborhood similarity loss or reconstruction loss is used, SSVH cannot achieve the best performance. Instead, the best performance is obtained when $\lambda=10^{-3}$, which indicates that it is necessary to use both information and the neighborhood structure loss contributes more to the performance.
	
	\subsubsection{Effect of K1 and K2 in Neighborhood structure}

	In this subsection, we evaluate mAP@K of different combinations of $K1$ and $K2$ on the FCVID dataset.  We tune both $K2$ and $K1$ from $5$, $10$, $20$, $30$, $40$, $50$, and the results are shown in Tab. \ref{tab.K1_K2}. When $K1$ or $K2$ is relatively high (e.g.,50 ), the worst performance is achieved. When both $K1$ and $K2$ are set as $10$, best performance is achieved. If $K1$ or $K2$ is set to a small number, there are not enough neighboring information to preserve; while if $K1$ or $K2$ is set to a large number, then similar videos categories will be considered as one category.

\eat{
	And we found that K1 as 10 and K2 as 10 will outperform other options. It is obviously to see that different K1 and K2 will result in different performances. If K1 and K2 are too small, neighbor structure will preserve less similarity information and level off to the performance without neighbor structure gradually. Also similar category videos will also be seen as one category if K1 and K2 are much large, which will also have an negative influence on final results. Besides, with K1 and K2 increases gradually, the performance with various K2 changes more significantly. It means that K2 play a decisive role of preserving similarity information which verify our initial guess that it is wise to add K2 in neighbor structure rather than only increasing K1.
}

	\subsubsection{Comparison with different approximate activation functions}
	{We have different ways to approximate sgn function. One is $p(\textbf{h})$ as we illustrate in Sec.\ref{optim} and the another is $tanh$. In this subsection, we compare these two activation functions on FCVID dataset, and the experimental results are shown in Tab.\ref{appro}. Note that the mAP is calculated using top20 retrieval results. From Tab.\ref{appro}, both $tanh(\textbf{h})$ and $p(\textbf{h})$ get satisfactory performance. And $p(\textbf{h})$ performs slightly better than $tanh(\textbf{h})$.}
	
	\begin{table}[]
		\small
		\centering
		\caption{{The result comparison of different activation functions on FCVID dataset. $p(h)$ denotes approximated sgn function defined in Eq.\ref{eq.ph}, $tanh(h)$ denotes tanh function. mAP is calculated using top20 retrieval results.}}
		\label{appro}
		\begin{tabular}{c|c c c}
			\hline
			   & 64bits  & 128bits & 256bits \\ \hline
			$tanh$(\textbf{h}) & 24.87\%    & 31.58\% & 35.84\% \\ 
			$p(\textbf{h})$  & 26.12\% & 33.65\% & 39.25\% \\ 
		\end{tabular}
	\end{table}
	
	\subsubsection{Cross-dataset evaluation comparison}
	{Tab.\ref{tab.cross} lists the cross-dataset performance of all the hashing methods following~\cite{zhang2016play}. We can observe that all the methods suffer a performance drop when training on FCVID and testing on YFCC. This indicates that the performance is related to the scale of training dataset. When the scale of training dataset decreases, the mAP will drop accordingly. Domain shift is another possible reason for the performance decrease. 
	On the other hand, when we train on YFCC and test on FCVID, the mAP is improved significantly than that training and testing on FCVID. This demonstrate that for unsupervised hashing models, more training data is beneficial for performance gain, even though they are from different domains.}

	\begin{table}[]
		\footnotesize
		\centering
		\caption{{Cross-dataset mAP gain (\%) by Hamming ranking of 256 bits. mAP is calculated using top20 retrieval results.}}
		\label{tab.cross}
		\begin{tabular}{c|c|c|c|c|c|c}
			\hline
			mAP-256bits                                                     & Submod & MFH   & ITQ   & DH    & SSTH  & SSVH   \\ \hline
			\begin{tabular}[c]{@{}c@{}}train:FCVID\\ test:YFCC\end{tabular} & -33.8$\textcolor{red}{\downarrow}$  & -24.7$\textcolor{red}{\downarrow}$ & -4.76$\textcolor{red}{\downarrow}$ & \textbf{-2.04 $\textcolor{red}{\downarrow}$ } & -11.6$\textcolor{red}{\downarrow}$ & -19.51$\textcolor{red}{\downarrow}$ \\ \hline
			\begin{tabular}[c]{@{}c@{}}train:YFCC\\ test:FCVID\end{tabular} & -20.3$\textcolor{red}{\downarrow}$  & 2.38$\textcolor{green}{\uparrow}$  & -8.26$\textcolor{red}{\downarrow}$ & -3.93$\textcolor{red}{\downarrow}$ & 7.58$\textcolor{green}{\uparrow}$  & \textbf{8.45$\textcolor{green}{\uparrow}$}   \\ \hline
		\end{tabular}
	\end{table}
	
	\subsubsection{Comparison with State-Of-The-Arts on FCVID}
	
	Figure \ref{fig.alg.hash} shows the comparison of our SSVH with several state-of-the-art video hashing methods.
	1) Obviously, SSVH achieves the best performance at all bit lengths on the FCVID dataset. Specifically, it outperforms the state-of-the-art method SSTH by 9.6\%, 9.3\% 9.6\%, 9.4\% and 9.4\% for mAP@K (K=20,40,60,80,100) when the code length is 256. 
	2) The advantage of our SSVH is obvious when the code length is relatively large, e.g., 32, 64, 128 and 256 bits.
	However, our method does not show the superiority over SSTH on short codes such as 8 and 16 bits, even though it performs better than the other baselines. One possible reason is that short codes carry too less information for reconstructing the video content, and preserving the neighborhood structure.
	3) In general, with the increase of code length, the mAP increases as well. For example, the mAP@20 for our method increases from 16.2\% with code length of 8, to 37.9\% with the code length of 256. This indicates that the code length plays an important role in video retrieval, and our method is suitable for longer hash codes.

	\subsubsection{Comparison with State-Of-The-Arts on YFCC}
	For the YFCC, the proposed SSVH consistently outperforms the other methods as shown in Fig.\ref{fig.alg.hash}. 1) SSVH shows significantly better performance compared with the baselines (ITQ, MFH, Submod, DH and SSTH) for different bits. Specifically, compared to the best counterpart SSTH, the performance is improved by 13\% in average in terms of mAP for 256 bits on YFCC. 
	2) An interesting phenomenon is that the performance improvement of SSVH is more significant for short codes such as 8 or 16 bits, which is different from the FCVID dataset. 
	3) Similar to the results on the FCVID dataset, the mAP increases for with the increase of code length. SSTH is a strong competitor when the code length is 64, 128 and 256. The performance gap of between our method and SSTH becomes marginal in terms of mAP@60, mAP@80 and mAP@100. Because the training strategy of YFCC is different from FCVID and we ignore lots of similarity information of video pairs due to the limitation of dataset scale. 
	
    \subsection{Qualitative Results}
    The qualitative results are shown in Fig. \ref{fig.performance}. The left results are obtained from the FCVID dataset, while the right results are obtained from the YFCC dataset. In this sub-experiments, both SSTH and SSVH generate 256 bits hash codes. In addition, videos marked with green indicate correct results, while videos marked with red are wrong results. From  Fig. \ref{fig.performance}, we can see that in general, SSVH can obtain better results. Given two queries ``Guitar Performance'' and ``Ski Slope'', both SSTH and SSVH obtain correct top5 retrieval videos. This indicates that modeling temporal information is beneficial to discriminate concepts that involve human actions.  More interestingly, we can observe that SSVH consistently outperforms the SSTH on both FCVID and YFCC datasets. This indicates that SSVH is able to obtain better temporal information for video hashing than SSTH.  As another example illustrated in Fig. \ref{fig.performance} (i.e., ``Tornado'' in FCVID and ``Patio'' in YFCC), it seems that capturing visual appearances is sufficient for retrieving them. This indicates that both SSVH and SSTH are also powerful for video categories that are not likely to be distinguished by temporal information.
    We also illustrated some failure cases for both methods. Two examples is the food related events like ``Making Hotdog'' in FCVID dataset and ``Pantry'' in YFCC dataset. Both methods cannot distinguish these actions from similar videos e.g., ``nail paining''.

\section{Conclusion}
{In this paper, we have extend the novel unsupervised deep hashing method (SSTH), named self-supervised video hashing (SSVH). To the best of our knowledge, SSVH is the first method which learns the video hash codes by simultaneous reconstructing the video contents and neighborhood structure. Experiments on real dataset show that SSVH can significantly outperform the others and achieve the state-of-the-art performance for video retrieval. {However, we also find some shortcomings of our model. We use pre-trained VGGNet to extract video frame features which ignore consecutiveness and temporal information of videos. In the future, we will consider extracting motion feature to our model and fusing multiple features to improve video retrieval performance.}}

\section{Acknowledgments}
\label{sec.ack}
This work is supported by the Fundamental Research Funds for the Central Universities (Grant No. ZYGX2016J085), the National Natural Science Foundation of China (Grant No. 61772116, No. 61502080, No. 61632007) and the 111 Project (Grant No. B17008).

\bibliographystyle{IEEEtran}

\begin{thebibliography}{10}
	\providecommand{\url}[1]{#1}
	\csname url@samestyle\endcsname
	\providecommand{\newblock}{\relax}
	\providecommand{\bibinfo}[2]{#2}
	\providecommand{\BIBentrySTDinterwordspacing}{\spaceskip=0pt\relax}
	\providecommand{\BIBentryALTinterwordstretchfactor}{4}
	\providecommand{\BIBentryALTinterwordspacing}{\spaceskip=\fontdimen2\font plus
		\BIBentryALTinterwordstretchfactor\fontdimen3\font minus
		\fontdimen4\font\relax}
	\providecommand{\BIBforeignlanguage}[2]{{%
			\expandafter\ifx\csname l@#1\endcsname\relax
			\typeout{** WARNING: IEEEtran.bst: No hyphenation pattern has been}%
			\typeout{** loaded for the language `#1'. Using the pattern for}%
			\typeout{** the default language instead.}%
			\else
			\language=\csname l@#1\endcsname
			\fi
			#2}}
	\providecommand{\BIBdecl}{\relax}
	\BIBdecl
	
	\bibitem{zhang2016play}
	H.~Zhang, M.~Wang, R.~Hong, and T.-S. Chua, ``Play and rewind: Optimizing
	binary representations of videos by self-supervised temporal hashing,'' in
	\emph{Proceedings of the 2016 ACM on Multimedia Conference}.\hskip 1em plus
	0.5em minus 0.4em\relax ACM, 2016, pp. 781--790.
	
	\bibitem{ye2013large}
	G.~Ye, D.~Liu, J.~Wang, and S.-F. Chang, ``Large-scale video hashing via
	structure learning,'' in \emph{Proceedings of the IEEE International
		Conference on Computer Vision}, 2013, pp. 2272--2279.
	
	\bibitem{song2011multiple}
	J.~Song, Y.~Yang, Z.~Huang, H.~T. Shen, and R.~Hong, ``Multiple feature hashing
	for real-time large scale near-duplicate video retrieval,'' in
	\emph{Proceedings of the 19th ACM international conference on
		Multimedia}.\hskip 1em plus 0.5em minus 0.4em\relax ACM, 2011, pp. 423--432.
	
	\bibitem{song2017deep}
	J.~Song, T.~He, L.~Gao, X.~Xu, and H.~T. Shen, ``Deep region hashing for
	efficient large-scale instance search from images,'' \emph{arXiv preprint
		arXiv:1701.07901}, 2017.
	
	\bibitem{DBLP:journals/pami/GongLGP13}
	\BIBentryALTinterwordspacing
	Y.~Gong, S.~Lazebnik, A.~Gordo, and F.~Perronnin, ``Iterative quantization: {A}
	procrustean approach to learning binary codes for large-scale image
	retrieval,'' \emph{{IEEE} Trans. Pattern Anal. Mach. Intell.}, vol.~35,
	no.~12, pp. 2916--2929, 2013. [Online]. Available:
	\url{http://dx.doi.org/10.1109/TPAMI.2012.193}
	\BIBentrySTDinterwordspacing
	
	\bibitem{LiuWJJC12}
	W.~Liu, J.~Wang, R.~Ji, Y.~Jiang, and S.~Chang, ``Supervised hashing with
	kernels,'' in \emph{CVPR}, 2012.
	
	\bibitem{smeulders2000content}
	A.~W. Smeulders, M.~Worring, S.~Santini, A.~Gupta, and R.~Jain, ``Content-based
	image retrieval at the end of the early years,'' \emph{IEEE Transactions on
		pattern analysis and machine intelligence}, vol.~22, no.~12, pp. 1349--1380,
	2000.
	
	\bibitem{datta2008image}
	R.~Datta, D.~Joshi, J.~Li, and J.~Z. Wang, ``Image retrieval: Ideas,
	influences, and trends of the new age,'' \emph{ACM Computing Surveys (Csur)},
	vol.~40, no.~2, p.~5, 2008.
	
	\bibitem{wang2012semi}
	J.~Wang, S.~Kumar, and S.-F. Chang, ``Semi-supervised hashing for large-scale
	search,'' \emph{IEEE Transactions on Pattern Analysis and Machine
		Intelligence}, vol.~34, no.~12, pp. 2393--2406, 2012.
	
	\bibitem{cao2012submodular}
	L.~Cao, Z.~Li, Y.~Mu, and S.-F. Chang, ``Submodular video hashing: a unified
	framework towards video pooling and indexing,'' in \emph{Proceedings of the
		20th ACM international conference on Multimedia}.\hskip 1em plus 0.5em minus
	0.4em\relax ACM, 2012, pp. 299--308.
	
	\bibitem{wu2016harnessing}
	Z.~Wu, Y.~Fu, Y.-G. Jiang, and L.~Sigal, ``Harnessing object and scene
	semantics for large-scale video understanding,'' in \emph{Proceedings of the
		IEEE Conference on Computer Vision and Pattern Recognition}, 2016, pp.
	3112--3121.
	
	\bibitem{wang2013action}
	H.~Wang and C.~Schmid, ``Action recognition with improved trajectories,'' in
	\emph{Proceedings of the IEEE international conference on computer vision},
	2013, pp. 3551--3558.
	
	\bibitem{wang2016learning}
	J.~Wang, W.~Liu, S.~Kumar, and S.-F. Chang, ``Learning to hash for indexing big
	data—a survey,'' \emph{Proceedings of the IEEE}, vol. 104, no.~1, pp.
	34--57, 2016.
	
	\bibitem{donahue2015long}
	J.~Donahue, L.~Anne~Hendricks, S.~Guadarrama, M.~Rohrbach, S.~Venugopalan,
	K.~Saenko, and T.~Darrell, ``Long-term recurrent convolutional networks for
	visual recognition and description,'' in \emph{Proceedings of the IEEE
		conference on computer vision and pattern recognition}, 2015, pp. 2625--2634.
	
	\bibitem{krizhevsky2012imagenet}
	A.~Krizhevsky, I.~Sutskever, and G.~E. Hinton, ``Imagenet classification with
	deep convolutional neural networks,'' in \emph{Advances in neural information
		processing systems}, 2012, pp. 1097--1105.
	
	\bibitem{girshick2015fast}
	R.~Girshick, ``Fast r-cnn,'' in \emph{Proceedings of the IEEE International
		Conference on Computer Vision}, 2015, pp. 1440--1448.
	
	\bibitem{you2016image}
	Q.~You, H.~Jin, Z.~Wang, C.~Fang, and J.~Luo, ``Image captioning with semantic
	attention,'' in \emph{Proceedings of the IEEE Conference on Computer Vision
		and Pattern Recognition}, 2016, pp. 4651--4659.
	
	\bibitem{liong2016deep}
	V.~E. Liong, J.~Lu, Y.-P. Tan, and J.~Zhou, ``Deep video hashing,'' \emph{IEEE
		Transactions on Multimedia}, 2016.
	
	\bibitem{simonyan2014very}
	K.~Simonyan and A.~Zisserman, ``Very deep convolutional networks for
	large-scale image recognition,'' \emph{arXiv preprint arXiv:1409.1556}, 2014.
	
	\bibitem{he2016deep}
	K.~He, X.~Zhang, S.~Ren, and J.~Sun, ``Deep residual learning for image
	recognition,'' in \emph{Proceedings of the IEEE Conference on Computer Vision
		and Pattern Recognition}, 2016, pp. 770--778.
	
	\bibitem{venugopalan2015sequence}
	S.~Venugopalan, M.~Rohrbach, J.~Donahue, R.~Mooney, T.~Darrell, and K.~Saenko,
	``Sequence to sequence-video to text,'' in \emph{Proceedings of the IEEE
		International Conference on Computer Vision}, 2015, pp. 4534--4542.
	
	\bibitem{pan2016hierarchical}
	P.~Pan, Z.~Xu, Y.~Yang, F.~Wu, and Y.~Zhuang, ``Hierarchical recurrent neural
	encoder for video representation with application to captioning,'' in
	\emph{Proceedings of the IEEE Conference on Computer Vision and Pattern
		Recognition}, 2016, pp. 1029--1038.
	
	\bibitem{gu2016supervised}
	Y.~Gu, C.~Ma, and J.~Yang, ``Supervised recurrent hashing for large scale video
	retrieval,'' in \emph{Proceedings of the 2016 ACM on Multimedia
		Conference}.\hskip 1em plus 0.5em minus 0.4em\relax ACM, 2016, pp. 272--276.
	
	\bibitem{hochreiter1997long}
	S.~Hochreiter and J.~Schmidhuber, ``Long short-term memory,'' \emph{Neural
		computation}, vol.~9, no.~8, pp. 1735--1780, 1997.
	
	\bibitem{wang2017survey}
	J.~Wang, T.~Zhang, N.~Sebe, H.~T. Shen \emph{et~al.}, ``A survey on learning to
	hash,'' \emph{IEEE Transactions on Pattern Analysis and Machine
		Intelligence}, 2017.
	
	\bibitem{li2015feature}
	W.-J. Li, S.~Wang, and W.-C. Kang, ``Feature learning based deep supervised
	hashing with pairwise labels,'' \emph{arXiv preprint arXiv:1511.03855}, 2015.
	
	\bibitem{cao2017hashnet}
	Z.~Cao, M.~Long, J.~Wang, and P.~S. Yu, ``Hashnet: Deep learning to hash by
	continuation,'' \emph{arXiv preprint arXiv:1702.00758}, 2017.
	
	\bibitem{erin2015deep}
	V.~Erin~Liong, J.~Lu, G.~Wang, P.~Moulin, and J.~Zhou, ``Deep hashing for
	compact binary codes learning,'' in \emph{Proceedings of the IEEE Conference
		on Computer Vision and Pattern Recognition}, 2015, pp. 2475--2483.
	
	\bibitem{song2018quantization}
	J.~Song, L.~Gao, L.~Liu, X.~Zhu, and N.~Sebe, ``Quantization-based hashing: a
	general framework for scalable image and video retrieval,'' \emph{Pattern
		Recognition}, vol.~75, pp. 175--187, 2018.
	
	\bibitem{wang2016survey}
	J.~Wang, T.~Zhang, J.~Song, N.~Sebe, and H.~T. Shen, ``A survey on learning to
	hash,'' \emph{arXiv preprint arXiv:1606.00185}, 2016.
	
	\bibitem{revaud2013event}
	J.~Revaud, M.~Douze, C.~Schmid, and H.~J{\'e}gou, ``Event retrieval in large
	video collections with circulant temporal encoding,'' in \emph{Proceedings of
		the IEEE Conference on Computer Vision and Pattern Recognition}, 2013, pp.
	2459--2466.
	
	\bibitem{yue2015beyond}
	J.~Yue-Hei~Ng, M.~Hausknecht, S.~Vijayanarasimhan, O.~Vinyals, R.~Monga, and
	G.~Toderici, ``Beyond short snippets: Deep networks for video
	classification,'' in \emph{Proceedings of the IEEE conference on computer
		vision and pattern recognition}, 2015, pp. 4694--4702.
	
	\bibitem{sutskever2014sequence}
	I.~Sutskever, O.~Vinyals, and Q.~V. Le, ``Sequence to sequence learning with
	neural networks,'' in \emph{Advances in neural information processing
		systems}, 2014, pp. 3104--3112.
	
	\bibitem{song2017hierarchical}
	J.~Song, Z.~Guo, L.~Gao, W.~Liu, D.~Zhang, and H.~T. Shen, ``Hierarchical lstm
	with adjusted temporal attention for video captioning,'' \emph{arXiv preprint
		arXiv:1706.01231}, 2017.
	
	\bibitem{li2017adversarial}
	J.~Li, W.~Monroe, T.~Shi, A.~Ritter, and D.~Jurafsky, ``Adversarial learning
	for neural dialogue generation,'' \emph{arXiv preprint arXiv:1701.06547},
	2017.
	
	\bibitem{britz2017massive}
	D.~Britz, A.~Goldie, T.~Luong, and Q.~Le, ``Massive exploration of neural
	machine translation architectures,'' \emph{arXiv preprint arXiv:1703.03906},
	2017.
	
	\bibitem{kafle2016answer}
	K.~Kafle and C.~Kanan, ``Answer-type prediction for visual question
	answering,'' in \emph{Proceedings of the IEEE Conference on Computer Vision
		and Pattern Recognition}, 2016, pp. 4976--4984.
	
	\bibitem{yang2016stacked}
	Z.~Yang, X.~He, J.~Gao, L.~Deng, and A.~Smola, ``Stacked attention networks for
	image question answering,'' in \emph{Proceedings of the IEEE Conference on
		Computer Vision and Pattern Recognition}, 2016, pp. 21--29.
	
	\bibitem{VinyalsTBE15}
	O.~Vinyals, A.~Toshev, S.~Bengio, and D.~Erhan, ``Show and tell: {A} neural
	image caption generator,'' in \emph{{IEEE} Conference on Computer Vision and
		Pattern Recognition, {CVPR} 2015, Boston, MA, USA, June 7-12, 2015}, 2015,
	pp. 3156--3164.
	
	\bibitem{jamestheano}
	B.~James, B.~Olivier, B.~Fr{\'e}d{\'e}ric, L.~Pascal, and P.~Razvan, ``Theano:
	a cpu and gpu math expression compiler,'' in \emph{Proceedings of the Python
		for Scientific Computing Conference (SciPy)}.
	
	\bibitem{jiang2015exploiting}
	Y.-G. Jiang, Z.~Wu, J.~Wang, X.~Xue, and S.-F. Chang, ``Exploiting feature and
	class relationships in video categorization with regularized deep neural
	networks,'' \emph{arXiv preprint arXiv:1502.07209}, 2015.
	
	\bibitem{thomee2015new}
	B.~Thomee, D.~A. Shamma, G.~Friedland, B.~Elizalde, K.~Ni, D.~Poland, D.~Borth,
	and L.-J. Li, ``The new data and new challenges in multimedia research,''
	\emph{arXiv preprint arXiv:1503.01817}, vol.~1, no.~8, 2015.
	
	\bibitem{over2014trecvid}
	P.~Over, J.~Fiscus, G.~Sanders, D.~Joy, M.~Michel, G.~Awad, A.~Smeaton,
	W.~Kraaij, and G.~Qu{\'e}not, ``Trecvid 2014--an overview of the goals,
	tasks, data, evaluation mechanisms and metrics,'' in \emph{Proceedings of
		TRECVID}, 2014, p.~52.
	
\end{thebibliography}

\end{document}